\documentclass{article} 
\usepackage{iclr2027_conference,times}

\usepackage{amsmath,amsfonts,bm}

\def\eqref#1{equation~\ref{#1}}

\def\1{\bm{1}}

\DeclareMathAlphabet{\mathsfit}{\encodingdefault}{\sfdefault}{m}{sl}
\SetMathAlphabet{\mathsfit}{bold}{\encodingdefault}{\sfdefault}{bx}{n}

\usepackage{hyperref}
\usepackage{url}
\usepackage{graphicx}
\usepackage{booktabs}
\usepackage{float}
\usepackage{multirow}
\usepackage{amsmath}
\usepackage{cleveref}
\usepackage{wrapfig}
\usepackage{amssymb}
\usepackage{mathtools}
\usepackage{amsthm}
\usepackage{microtype}
\usepackage{subcaption}
\usepackage{multicol}
\usepackage{makecell}

\newcommand{\ours}{{ELROND}}

\title{\ours: Exploring and decomposing intrinsic
capabilities of diffusion models}

\author{%
  Paweł Skierś \\
  Warsaw University of Technology\\
  IDEAS Research Institute\\
  \texttt{pawel.skiers.dokt@pw.edu.pl} \\
  \And
  Emilia Kaczmarczyk \\
  University of Warsaw\\
  IDEAS Research Institute\\
  \And
  Tomasz Trzciński \\
  Warsaw University of Technology\\
  IDEAS Research Institute\\
  \And
  Kamil Deja \\
  Warsaw University of Technology\\
  IDEAS Research Institute\\
}
\author{%
  Paweł Skierś$^{1,2}$ \quad
  Emilia Kaczmarczyk$^{2,3}$ \quad
  Tomasz Trzciński$^{1,2}$ \quad
  Kamil Deja$^{1,2}$ \\[0.5em]
  $^{1}$Warsaw University of Technology \quad
  $^{2}$IDEAS Research Institute \quad
  $^{3}$University of Warsaw \\[0.3em]
  \texttt{pawel.skiers.dokt@pw.edu.pl}
}

\iclrfinalcopy 
\begin{document}

\maketitle

\begin{abstract}
A single text prompt passed to a diffusion model yields a wide range of visual outputs determined solely by a stochastic process, leaving users with no direct control over which semantic variations appear. Exploring this range is difficult: random search offers no guarantee of covering it, while prompt editing is coarse, as even a small change in wording can substantially alter the generated image. We argue that systematic exploration instead requires recovering how the model itself organizes the conditional distributions it can produce. We formalize this structure as a \emph{generative manifold}, and present \ours{}, a method for recovering its tangent space at a given conditioning. To that end, we collect gradients obtained by backpropagating the differences between stochastic realizations of a fixed prompt, and decompose them into interpretable directions using Principal Component Analysis or a Sparse Autoencoder. We show that our method recovers this subspace accurately in a controlled setting 
and validate it behaviorally on large-scale models. We also demonstrate that recovered structure enables broader exploration of the model's capabilities than methods relying on external representations, and mitigates mode collapse in distilled models without retraining.
\end{abstract}

\section{Introduction}
\label{sec:intro}

Text-to-image diffusion models~\citep{rombach2022highresolutionimagesynthesislatent,
podell2023sdxl} can generate a remarkably diverse set of images from a
single prompt. A prompt such as \emph{``An image of an elf''}
corresponds to anything from a Christmas creature to a majestic, immortal being, with the realized image determined entirely by the stochasticity. 
When a desired output is known in advance, existing methods can often narrow the generation through additional conditioning~\citep{zhang2023adding}, attention-based editing~\citep{hertz2022prompt0to0prompt}, or learned attribute directions~\citep{gandikota2024concept, baumann2025continuous}.
Creative exploration starts without such a target: users want to discover what the model can produce and then explore interesting variations further.
Prompt engineering therefore offers limited help and random resampling provides no systematic way to cover the model’s capabilities.
What is missing is a structure organizing the variations the model can produce.

\begin{figure}[t!]
    \centering
    \includegraphics[width=.8\linewidth]{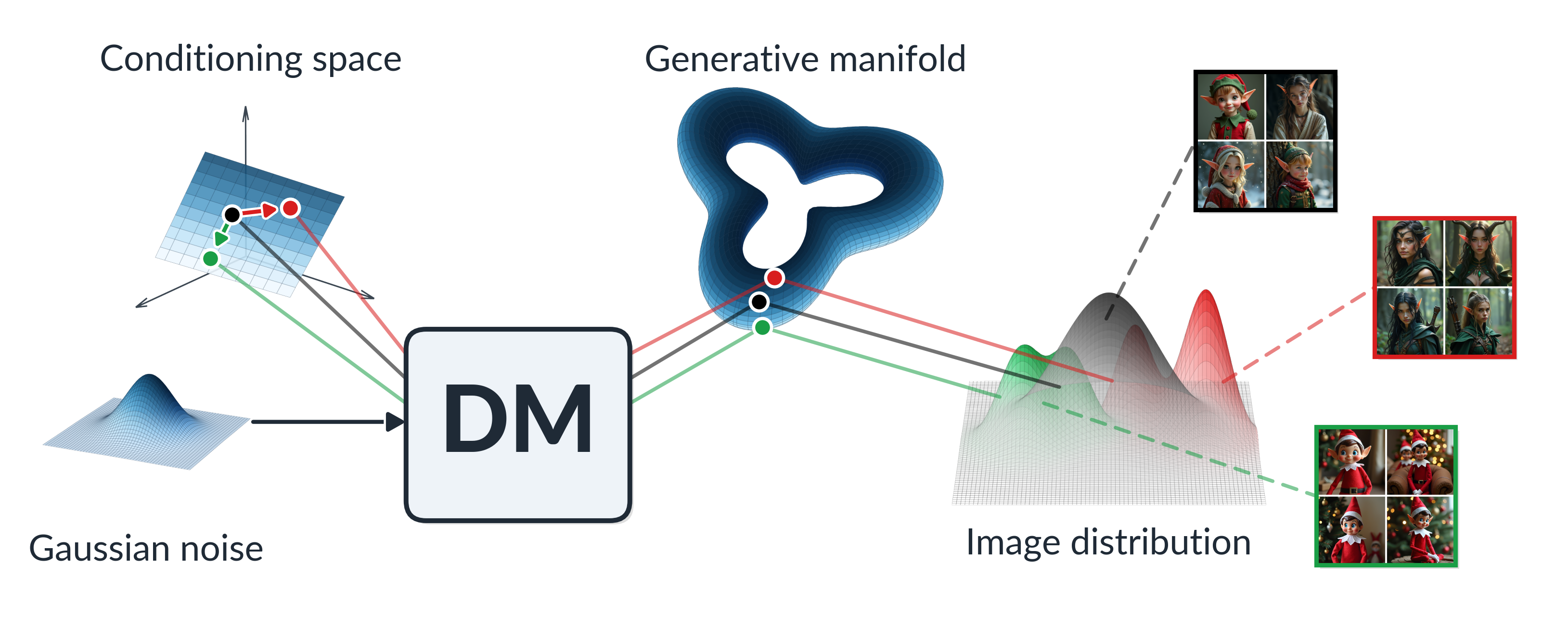}
    
    \caption{\textbf{Generative manifold of a diffusion model.}
    A conditioning $c$ selects a conditional image distribution $p_\theta(x\mid c)$
    on the \emph{generative manifold}, while Gaussian noise determines the
    individual realization sampled from that distribution. Nearby conditionings
    induce local movements on the manifold. \ours{} recovers its tangent space in
    the conditioning space.}
    \vspace{-1.0em}
    \label{fig:teaser_decomposition}
\end{figure}

To see what such a structure should represent, consider the generation
process itself. A diffusion model does not map a conditioning $c$ directly to a
single image but to a
distribution $p_\theta(x\mid c)$, from which an individual image is sampled. 
Thus, conditioning determines \emph{which distribution} is selected, while stochasticity determines \emph{which realization} is sampled.
For exploration, this distinction is important. Individual
images are the objects we observe, but the distribution is the object the
model controls. 
Exploring the model then means understanding how these distributions change as the conditioning changes.
We therefore view
the model through the family of conditional distributions it can realize (\cref{fig:teaser_decomposition}). We refer to this family as the model's \emph{generative manifold}, where each point is a distribution of
images, and samples from a fixed conditioning are realizations from that point. 
Access to this structure would make the exploration of a model's capabilities
systematic allowing moving between distributions 
and, in principle, reaching every single one the model is able to produce. 
Although the manifold itself is not directly accessible, the conditioning space provides coordinates for it, since small changes in the conditioning
induce local changes in the corresponding distribution. 


Recovering semantic structure in such spaces is the established purpose of mechanistic interpretability tools, most notably sparse autoencoders and linear probes.
Applying them here, however, exposes a fundamental limitation. These tools operate on activations gathered over diverse inputs and rely on the variance across them. 
Our setting admits none: the conditioning is fixed by construction, as the distribution available at a fixed prompt is exactly what we aim to characterize. Yet at a fixed prompt we have access to a different source of variation: stochasticity.

\ours{} exploits this observation by extracting the local signal directly from visual variation within a single conditional distribution.
Given two images generated from the same prompt under different random initializations, we construct a reconstruction objective that encourages the model's expected output to move from the characteristics of one realization toward those of the other. We then backpropagate this objective to the conditioning embedding, obtaining a direction in the conditioning space that corresponds to a local displacement on the generative manifold. 
Repeating this procedure over many pairs of samples provides a collection of local directions that span the tangent space at the distribution induced by the prompt.

We apply this construction to individual concepts 
by holding the remaining conditioning fixed and varying only the embedding of the
token representing the concept. 
We then decompose the recovered directions spanning the corresponding tangent space using Principal Component Analysis (PCA)
or Sparse Autoencoders (SAEs)~\citep{Olshausen1997SparseCW}, obtaining
interpretable directions that can be injected directly into the token
embedding during sampling.

We validate our framework first in a controlled setting where the tangent
space is known, showing that the recovered gradients recover the
ground-truth subspace. On large-scale text-to-image models, we show that the
model's sensitivity to its conditioning is concentrated in the recovered
subspaces. We then demonstrate the practical value of the
recovered structure for exploration, where our directions expose a wider range
of concept-relevant variation while maintaining semantic consistency, and for
mitigating mode collapse in distilled models. Our contributions can be summarized as follows: 

\begin{itemize}
    \item We introduce \ours{}, a method for recovering and decomposing the tangent space of the
    generative manifold that a diffusion model has learned.
    \item We validate the recovery both against a known ground truth in a controlled setting and behaviorally on large-scale models, where we show that the recovered structure enables exploration of a model's capabilities beyond what existing approaches reach.
    \item We quantitatively and qualitatively evaluate how we can benefit from our approach in practical task of mitigating mode collapse in distilled models.
\end{itemize}

\section{Related Work}
\paragraph{Precise generation in text-to-image diffusion models}
Achieving desired control over generated images necessitates mechanisms that go beyond text prompting. 
Standard approaches introduce additional conditioning by extending the model through adapters (ControlNet~\citep{zhang2023adding}, T2I-Adapter~\citep{DBLP:conf/aaai/MouWXW0QS24}). Without affecting the structure of the model architecture, \cite{rodriguez2024controlling,rodriguez2025lineas} have introduced a framework enabling fine-grained control by transporting the activations during inference.
Alternatively, attention-based methods like Prompt-to-prompt~\citep{hertz2022prompt0to0prompt} or Ledits++~\citep{brack2023ledits000} leverage cross-attention maps of a reference generation to localize changes, while~\cite{mokady2022null0text} optimizes a text-conditioning space. Most relevant to our proposed methodology, \cite{baumann2025continuous} identifies semantic directions within the CLIP text embeddings by backpropagating the difference in noise predictions between contrastive prompts (e.g \textit{"young person"} vs. \textit{"old person"}). However, this reliance on manually constructing contrastive pairs necessitates \textit{a priori} knowledge of the target attributes, rendering it infeasible for the unsupervised discovery of unknown semantic variations. 

\paragraph{Unsupervised concept discovery}
Without supervision, \citet{kwon2022diffusion} identify that U-net's bottleneck features possess a semantic structure, while \citet{dalva2024noiseclr} employs contrastive learning on the directions added to the intermediate diffusion steps, to identify steerable directions. Similarly, \citet{liu2023unsupervised} decompose an image into a set of randomly initialized concepts that can sum up into the observed composition by the summation of individual score functions. On the other hand, building on the model adaptation techniques, \citet{gandikota2024concept,gandikota2025sliderspace} automate the discovery process by applying PCA to CLIP embeddings of generated images, obtaining controllable basis vectors for a given concept without manual attribute specification. 
From a conceptually different direction, \citet{dravid2024interpreting} interpret the semantic directions in the adapter's weight space.

\paragraph{Sparse Autoencoders in Diffusion Models}
Recent research has successfully adapted Sparse Autoencoders (SAE) to interpret text-to-image diffusion models. Works by~\citet{cywinski2025saeuron, surkov2024one, huang2025tide} demonstrate that SAEs can decompose activations of diffusion models into interpretable features, allowing for precise steering of the generative process. In parallel, some works apply SAEs to interpret CLIP embeddings \citep{pach2025sparse, dreyer2025attributing} with additional applications when this model is used as conditioning for diffusion models \citep{kim2024textit, ye2025all}, including precise editing as shown by~\citet{kamenetsky2025saedit}. 

\paragraph{Geometry of learned representations}
The linear representation hypothesis holds that features are encoded as directions in activation space~\citep{park2023linear}, an assumption that underlies the use of sparse autoencoders for their recovery. Recent work argues that this picture is incomplete: \citet{engels2025not} identify irreducible multi-dimensional features, while \citet{modell2025origins} develop a theory of features encoded as manifolds and show that cosine similarity in representation space reflects their intrinsic geometry. 

\begin{figure*}[t] 
    \centering
    \includegraphics[width=\textwidth]{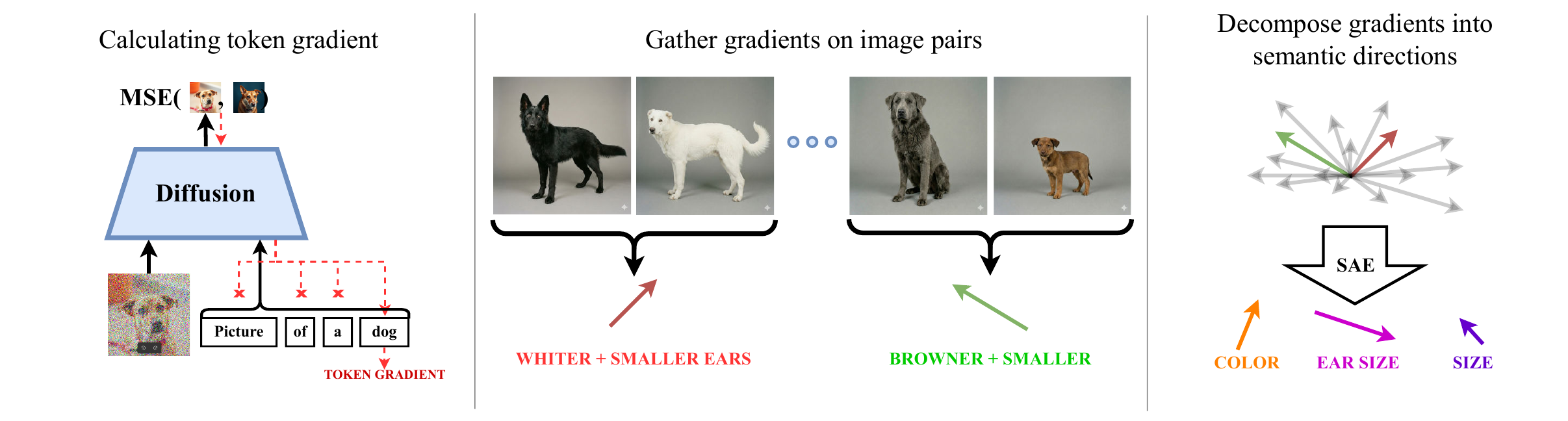}
    \vspace{-2em}
    \caption{
    To extract the latent directions for a given concept, we backpropagate the differences between image instances generated from the same prompt to the given token. We then disentangle the latent directions by training SAE / computing PCA on the gradients.
    \vspace{-1.0em}
    }
    \label{fig:method}
\end{figure*}

\section{Method}
\label{sec:method}
In this section, we introduce our framework. We first describe how to discover the distribution manifold by recovering its tangent space directly from sampled images and how to isolate this structure for individual concepts. We then explain how to decompose these recovered spaces into interpretable semantic directions, and how to use them for steering.
\subsection{Discovering the manifold}
\label{subsec:manifold}

A text-to-image diffusion model maps a text conditioning matrix $c \in \mathbb{R}^{L \times d}$, representing a sequence of $L$ tokens, to an image probability distribution $p_\theta(x \mid c)$. Let
\[
\mathcal{V} = \{ c \in \mathbb{R}^{L \times d} \mid \mathbb{P}_{x \sim
p_\theta(\cdot \mid c)}[x \in \mathcal{M}_{img}] \geq 1 - \delta \}
\]
be the set of conditionings that yield natural images, where $\mathcal{M}_{img}$ denotes the manifold of valid images. The family of distributions available to the model is then
\[
\mathcal{M} = \{ p_\theta(\cdot \mid c) \mid c \in \mathcal{V} \},
\]
which, following the manifold hypothesis, we posit to be a low-dimensional manifold. Our goal is to characterize the neighbourhood of one of its points, that is, to recover the tangent space $T_{\Phi(c)}\mathcal{M}$ at the distribution induced by a fixed prompt, where $\Phi(c) = p_\theta(\cdot \mid c)$.

The points of $\mathcal{M}$ are distributions, which we cannot manipulate directly. We therefore use the parametrization $\Phi$ of $\mathcal{M}$ and work in the conditioning space instead. Since $T_{\Phi(c)}\mathcal{M} = \operatorname{im}(d\Phi_c)$ and $d\Phi_c$ is injective on $\ker(d\Phi_c)^\perp$, this subspace is in bijection with the tangent space. Recovering $T_{\Phi(c)}\mathcal{M}$ is thus equivalent to finding its representation in the conditioning space
\[
\mathcal{T}_c = \ker(d\Phi_c)^\perp \subset \mathbb{R}^{L \times d}.
\]
This subspace contains every direction in the conditioning space that changes the images the prompt produces. For \emph{``An image of an elf''}, one such direction might shift the generated elves toward Christmas helpers, another toward armored warriors. Finding $\mathcal{T}_c$ therefore amounts to collecting
enough such directions to span it. We propose to recover them from the sampled images. 

Given a prompt with conditioning $c$, we first sample a dataset of $N$ images,
$\mathcal{X} = \{x^{(1)}, \dots, x^{(N)}\}$. To extract a single semantic direction, we sample a pair of distinct images
$(x^{(i)}, x^{(j)})$ from $\mathcal{X}$. We encode $x^{(i)}$ into the latent
space to obtain $z_0^{(i)}$ and apply forward diffusion to a noise level $t$:
\[z_t^{(i)} = \alpha_t z_0^{(i)} + \sigma_t \epsilon, \quad \epsilon \sim \mathcal{N}(0, I).\]

At any timestep $t$, the network $\epsilon_\theta$ predicts an estimate of the
clean latent, $\hat{z}_0$. By Tweedie's formula, this prediction serves as an
estimator for the conditional expectation of the clean image:
\[f_t(c, z_t^{(i)}) := \hat{z}_0(z_t^{(i)}, t, c) = \frac{z_t^{(i)} - \sigma_t \epsilon_\theta(z_t^{(i)}, t, c)}{\alpha_t} \approx \mathbb{E}[z_0 \mid z_t^{(i)}, c].\]
For a fixed $z_t^{(i)}$ and noise level $t$, the function $f_t$ defines a
parameterized smooth mapping from the conditioning into the visual latent space.

Because $f_t$ approximates the expected clean image, we can formulate an
objective to steer this expectation toward a target visual state. We compute the
standard diffusion reconstruction loss (MSE) between the prediction and the
latent representation of the second image $x^{(j)}$:
\[\mathcal{L}_{sem} = \|f_t(c, z_t^{(i)}) - z_0^{(j)}\|_2^2\]
(see App.~\ref{app:loss_ablation} for alternative formulations), and
backpropagate it to obtain the gradient with respect to the conditioning,
$g_{i,j} = \nabla_{c} \mathcal{L}_{sem}$.

Intuitively, minimizing this loss forces the model to alter the conditioning
such that the expected generated output shifts from the characteristics of
$x^{(i)}$ toward $x^{(j)}$. Formally, let $J_{c,t} = \nabla_{c} f_t(c,
z_t^{(i)})$ be the Jacobian matrix of the denoising mapping. Applying the chain
rule to the semantic loss yields
\[g_{i,j} = 2 J_{c,t}^\top (f_t(c, z_t^{(i)}) - z_0^{(j)}).\]
The visual difference vector $(f_t(c, z_t^{(i)}) - z_0^{(j)})$ represents a
macroscopic transformation toward the target image. When multiplied by
$J_{c,t}^\top$, any component of this difference lying in the left null space of
the Jacobian - representing changes that the conditioning cannot control - is mapped to zero.
Consequently, $g_{i,j} \in \ker(J_{c,t})^\perp$, and since a displacement that
leaves $\Phi(c)$ unchanged also leaves every $f_t(\cdot, z_t)$ unchanged, we have
$\ker(d\Phi_c) \subseteq \ker(J_{c,t})$. Each gradient therefore lies in the
tangent space we set out to recover, $g_{i,j} \in \mathcal{T}_c$.

Repeating this process for many pairs gives a dataset of gradients $\mathcal{G} = \{g_1, \dots, g_M\}$, representing the semantic directions that span the tangent space of the concept in the embedding space.

\subsection{Restricting to a Concept}
\label{subsec:restriction}

The construction above recovers the variations of the entire generated image. In practice, however, one is more often interested in something more specific, such as the variations of a single concept within a scene, while the rest of it stays intact. Prior work shows that the influence of individual tokens is largely localized, with cross-attention maps attributing image content to the corresponding words~\citep{hertz2022prompt0to0prompt, tang2023daam}. Following these observations, we assume that the characteristics of a concept are controlled predominantly by its own tokens.

Under this assumption, fixing the remaining tokens and varying only the embedding $e \in \mathbb{R}^d$ of the token of interest traces a submanifold of $\mathcal{M}$, which we refer to as the \emph{concept manifold}. Its tangent space is a subspace of $\mathcal{T}_c$ and, since $\nabla_e \mathcal{L}_{sem}$ is the block of $\nabla_c \mathcal{L}_{sem}$ corresponding to that token, the derivation above carries over unchanged: we collect gradients exactly as before, restricting the backward pass to the concept tokens, and obtain $\mathcal{G} \subset \mathcal{T}_e$.

\subsection{Discovering Latent Directions}
\label{subsec:latent_directions}
The raw gradients in $\mathcal{G}$ often contain entangled information about several semantic changes together with unimportant noise. To extract disentangled, semantic directions, we decompose the set of gradients with two approaches:

\textbf{Principal Component Analysis (PCA).} We perform PCA on $\mathcal{G}$. The resulting latent directions are defined as the principal components (eigenvectors) $v_k$. To filter out noise, we retain only components with explained variance above a threshold $\tau$. 

\textbf{Sparse Autoencoder Decomposition.} For more interpretable directions, we train a Top-k Sparse Autoencoder (SAE)~\citep{bussmann2024batchtopk} on the set of gradients $\mathcal{G}$. To that end, we use a model composed of two neural networks: an encoder and a decoder trained using a reconstruction loss with sparsity constraints. Given an input vector $v \in \mathbb{R}^d$, the encoder projects it to a higher-dimensional latent space $h \in \mathbb{R}^n$ using a linear transformation followed by a TopK activation function:
$h = \text{TopK}(W_{enc}(v - b_{pre})),$
where $W_{enc} \in \mathbb{R}^{n \times d}$ is the encoder weight matrix, $b_{pre} \in \mathbb{R}^d$ is a pre-encoder bias, and $\text{TopK}(\cdot)$ is a non-linearity that retains only the $k$ largest activation values, setting the rest to zero. This architectural constraint enforces a strict sparsity level of $k$ active features per input.
The decoder then reconstructs the input from the sparse latent code $h$, and minimizes the reconstruction error:
    \[\mathcal{L}_{SAE} = \|v - (W_{dec}h + b_{pre})\|_2^2,\]
where $W_{dec} \in \mathbb{R}^{d \times n}$ is the decoder weight matrix.
Post-training, we identify the set of ``alive'' latent features (with non-zero activation density). The decoder weights corresponding to these features, $d_k = (W_{dec})_{:,k}$, represent the disentangled semantic directions. The sparsity constraint ensures that these directions represent fundamental, primitive attributes of the concept.

\subsection{Semantic Steering}
Given a discovered latent direction $d$ (derived either from PCA or SAE), we can precisely steer the generation process to obtain a final generation with a characteristic defined by 
a given direction. For a given prompt, let $e$ be the original embedding of the concept token. We modulate the embedding by injecting the latent direction:
    $e' = e + \lambda \cdot d, $
where $\lambda \in \mathbb{R}$ is a strength parameter. The modified embedding is then passed to the diffusion model. This method scales naturally to complex prompts with multiple subjects, as directions are derived relative to specific tokens. Consequently, we can apply distinct directions to different subjects simultaneously with limited cross-concept interference.

\section{Experiments}
\label{sec:experiments}

For the large-scale experiments, we use SDXL~\citep{podell2023sdxl} and its distilled version SDXL-DMD~\citep{yin2024improved}, and show in App.~\ref{app:flux} that \ours{} generalizes to the transformer-based Flux Schnell~\citep{flux1schnell_hf}. The text conditioning acts as a concatenation of embeddings: from two distinct CLIP models for SDXL, and from a CLIP and T5 encoder for Flux-Schnell. Consequently, our method operates directly on this joint embedding space (dim=2048 for SDXL, dim=4096 for Flux). To analyze a specific concept $c$, we employ a generic prompt template $P = \text{``A picture of } \langle c \rangle \text{''}$. For each concept, we sample $N=10,000$ images using the base model and collect $10,000$ semantic gradients as described in Section~\ref{sec:method}. 
For the synthetic model, we collect this set independently at each diffusion timestep to fully characterize the recovered subspace. For the large-scale models, collecting
gradients at every timestep would be computationally prohibitive, and unnecessary at the same time, as the conditioning signal is mosty used at the beginning of the sampling sequence. So, we restrict the backward pass to the highest-noise timestep ($t=T$) (see App.~\ref{app:timesteps_selection} for ablation). We provide the used intervention strengths $\lambda$ and an ablation of its effect in App.~\ref{app:intervention_str}. For the PCA decomposition, we use a significance threshold of $\tau = 2.5 / d_{embed}$, and for SAE training $k = 32$ and $d = d_{embed}$ (see App.~\ref {app:sae_ablation} for ablations). The architecture and training details of the toy model used in Sec.~\ref{subsec:toy} are provided in App.~\ref{app:toy_model}.

\subsection{Validating the recovered subspace}

The validity of our method relies on the recovered subspace matching the concept's tangent space. Since $\mathcal{T}_c$ of a large-scale model is unknown, we first verify the recovery on an artificial dataset.

\label{subsec:toy}
\paragraph{Synthetic dataset validation} We train a diffusion model on a synthetic dataset of $64 \times 64$ images, each containing a single geometric shape with known attributes: size, color, and position. Each shape has a different set of attributes that change between instances. The model is conditioned on a vector $u \in \mathbb{R}^{3+A}$ mapped into 64-dimensional space by a linear encoder $E$. The first three coordinates of $u$ one-hot encode the shape. Each of the remaining $A$ coordinates, taking values in $[0,1]$, controls the distribution one attribute is drawn from rather than setting its value directly. At 0.5, the attribute is drawn from a wide window centered in the middle of its range; moving the coordinate toward either end slides and narrows that window, and near the extremes it closes entirely, fixing the attribute to a single value. This mirrors text conditioning, where \emph{``a cat''} admits every cat the model has learned and \emph{``a big cat''} restricts the distribution to a given subset. 
\begin{wraptable}{r}{0.45\linewidth}
    \centering
    \caption{\small Subspaces recovered on the synthetic dataset: dimension
    estimated by VAEase, and precision and recall of the PCA subspace with
    respect to the ground truth.}
    \label{tab:toy_results}
    \small
    \setlength{\tabcolsep}{4pt}
    \begin{tabular}{lccc}
        \toprule
        \textbf{Shape} & \textbf{Square} & \textbf{Triangle} & \textbf{Circle} \\
        \midrule
        Ground-truth dim. & 3 & 4 & 6 \\
        Estimated dim.    & 3 & 4 & 8 \\
        \midrule
        Recall            & 0.956 & 0.939 & 0.968 \\
        Precision         & 0.956 & 0.939 & 0.726 \\
        \bottomrule
    \end{tabular}
\end{wraptable}
We assume that the model has learned the dataset faithfully and, given any conditioning, generates shapes with correct distributions of attributes, which, given the simplicity of the dataset, we take to hold approximately. Consequently, the subspace $\mathcal{T}_c$ at a given shape $c$ is exactly the image under $E$ of the coordinate directions of its varying attributes.

We collect 10 000 gradients from each of 20 uniformly spaced timesteps described in Sec.~\ref{sec:method} and first estimate the dimensionality of the recovered subspace using VAEase~\citep{lu2025sparse}. As shown in
Tab.~\ref{tab:toy_results}, VAEase estimates only $3$-$8$ dimensions across the different shapes, indicating that the recovered variation is concentrated in a low-dimensional subspace. Moreover, the estimated dimensionalities closely match the ground-truth dimensions and preserve their relative ordering across shapes. The only discrepancy occurs for the circle, for which VAEase estimates $8$ dimensions compared to the $6$-dimensional ground truth. We attribute them to imperfections in model training; see App.~\ref{app:toy_discussion} for further discussion.

We next test whether these estimated dimensions correspond to the correct subspace. We perform PCA on the collected gradients and retain the top $\hat{d}_c$ components, where $\hat{d}_c$ is the VAEase estimate, yielding an orthonormal basis $\{u_i\}$ of the recovered subspace $\hat{\mathcal{T}_c}$. Letting $P$ and $\hat{P}$ denote the orthogonal projections onto the ground-truth subspace $\mathcal{T}_c$ and recovered subspace $\hat{\mathcal{T}_c}$, respectively, and $\{v_j\}$ an orthonormal basis of $\mathcal{T}_c$, we measure precision and recall defined as:
\[
\text{precision} = \frac{1}{\dim \hat{\mathcal{T}_c}} \sum_i \|P u_i\|^2,
\qquad
\text{recall} = \frac{1}{\dim \mathcal{T}_c} \sum_j \|\hat{P} v_j\|^2.
\]

Intuitively, recall measures how much of the ground-truth 
subspace is spanned by the recovered one, while precision measures how much of the recovered subspace lies within the ground-truth one. As shown in Tab.~\ref{tab:toy_results}, recall exceeds $90\%$ for all shapes, indicating that the recovered directions span the ground-truth tangent space well. Precision is likewise high, showing that the recovered subspace is predominantly concentrated within the correct tangent space. The only notable deviation is the circle, whose precision falls below $90\%$, consistent with the overestimate of its dimensionality by VAEase.

\paragraph{Validation on Large-Scale Models}
\label{subsec:large_scale_validation}
\begin{figure}[t]
    \centering
    \begin{minipage}[c]{0.62\linewidth}
        \centering
        \includegraphics[width=\linewidth]{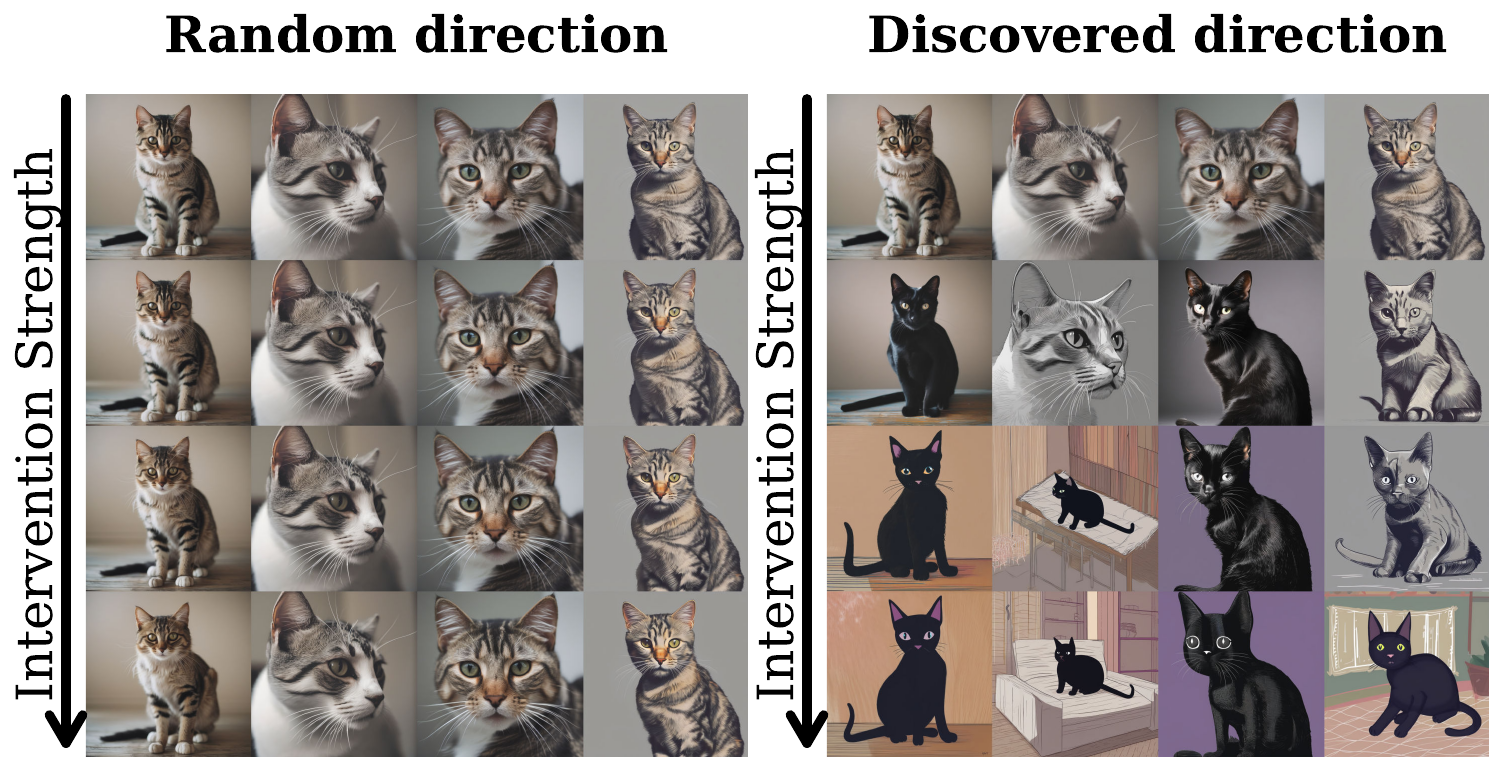}
        \vspace{-1.5em}
        \caption{\small The diffusion model responds far more strongly to
        directions found by \ours{} than to random directions of the same norm.}
        \label{fig:random_vs_found}
    \end{minipage}
    \hfill
    \begin{minipage}[c]{0.34\linewidth}
        \centering
        \captionof{table}{\small Mean DreamSim distance ($\uparrow$) between
        images generated before and after perturbing the embedding.}
        \label{tab:manifold_proof}
        \small
        \begin{tabular}{lcc}
            \toprule
            \textbf{Concept} & \textbf{Random} & \textbf{Discovered} \\
            \midrule
            Cat     & 0.068 & \textbf{0.230} \\
            Monster & 0.095 & \textbf{0.288} \\
            Car     & 0.085 & \textbf{0.249} \\
            Dog     & 0.073 & \textbf{0.223} \\
            Person  & 0.091 & \textbf{0.261} \\
            \bottomrule
        \end{tabular}
    \end{minipage}
\end{figure}

\begin{wrapfigure}{r}{0.39\linewidth}
        \centering
        \vspace{-1.0em}
        \includegraphics[width=\linewidth]{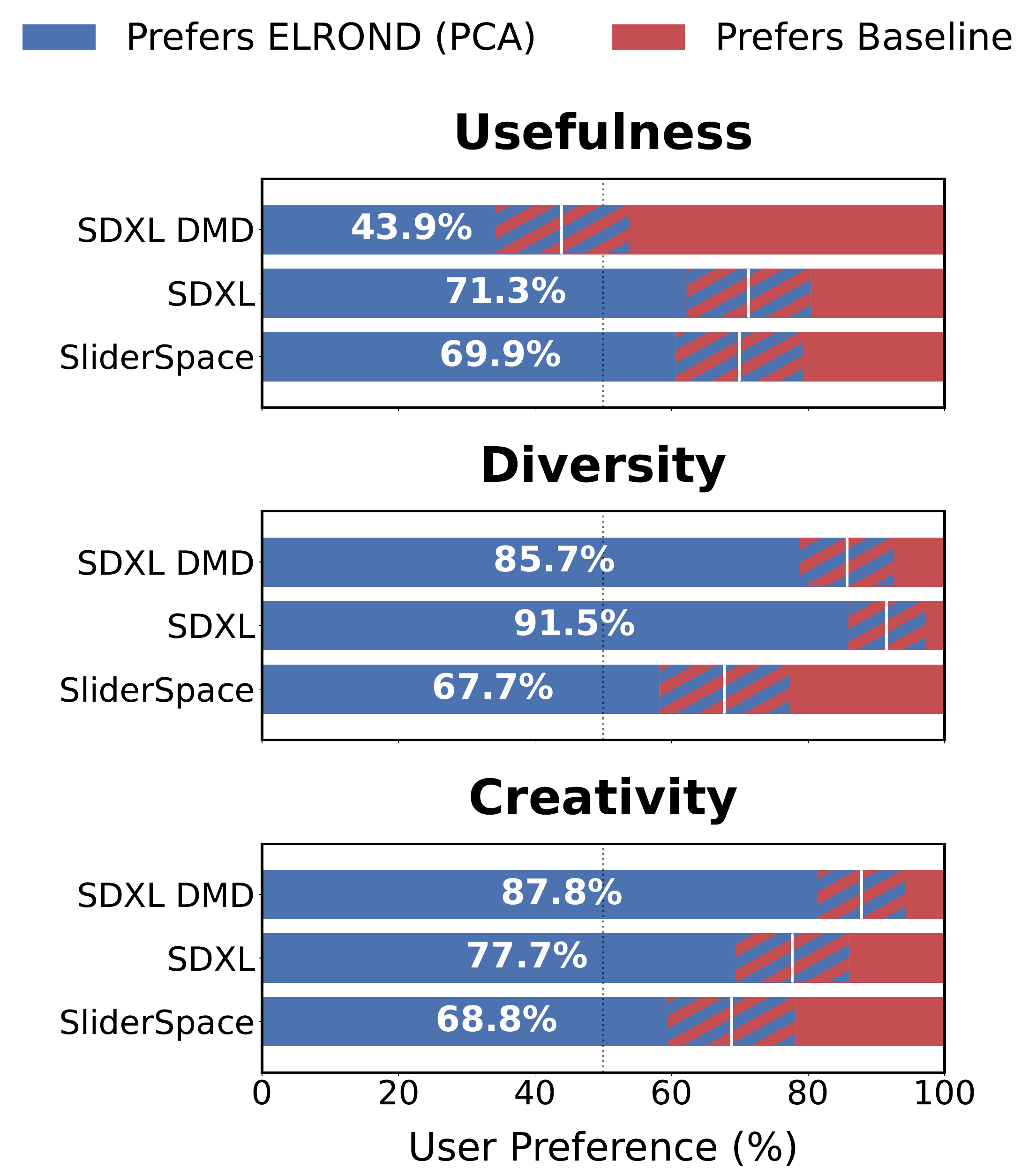}
        \caption{\small User study results evaluating ELROND against baselines across usefulness, diversity, and creativity.}
        \label{fig:elrond_vs}
        \vspace{-1.5em}
\end{wrapfigure}
On large-scale text-to-image models, no ground truth is available, and we therefore validate the recovery behaviourally. If the collected gradients span $\mathcal{T}_c$, the model's sensitivity to its conditioning should be concentrated there: displacements within the recovered subspace should change the generated images, while displacements of equal magnitude outside it should leave them unchanged, as the embedding space is high-dimensional and the manifold occupies a small part of it.

As illustrated in Fig.~\ref{fig:random_vs_found}, modulating the embedding with a discovered direction significantly alters the semantic content of the image, while adding a random vector of the same norm results in negligible visual change. To quantify this, we sample 2,000 image pairs per concept, generating a base image and a perturbed one for each, and measure the perceptual shift using DreamSim~\citep{fu2023dreamsim}. Tab.~\ref{tab:manifold_proof} reports that the mean distance under our directions is substantially higher than under random perturbations of equal magnitude.

\subsection{Exploring the Model's Capabilities}
\label{subsec:exploration}
\begin{table}
    \centering
    \caption{Decomposition of the measured variation. VLM ratings (1--5) of how much
    each factor differs within a pair, and the share of pairs attributed to each
    primary driver.}
    \label{tab:drift_decomposition}
    \vspace{-0.5em}
    \small
    \resizebox{\linewidth}{!}{
        \begin{tabular}{lcccccc}
            \toprule
            Method & Targ. attr. $\uparrow$ & Comp. $\downarrow$ & Backgr. $\downarrow$ & Style $\downarrow$ & Artif. $\downarrow$ & Primary driver $\uparrow$ \\
            \midrule
            \ours{}-PCA  & 4.76 & 1.89 & 1.87 & 1.84 & 1.27 & 96.7\% \\
            \ours{}-SAE  & \textbf{4.79} & \textbf{1.62} & \textbf{1.61} & \textbf{1.52} & \textbf{1.10} & \textbf{97.7\%} \\
            SliderSpace  & 4.67 & 1.68 & 1.65 & 1.55 & 1.25 & 94.0\%\\
            \bottomrule
        \end{tabular}
    }
    \vspace{-0.5em}
\end{table}

\begin{figure}[t!] 
    \centering
    \includegraphics[width=\textwidth]{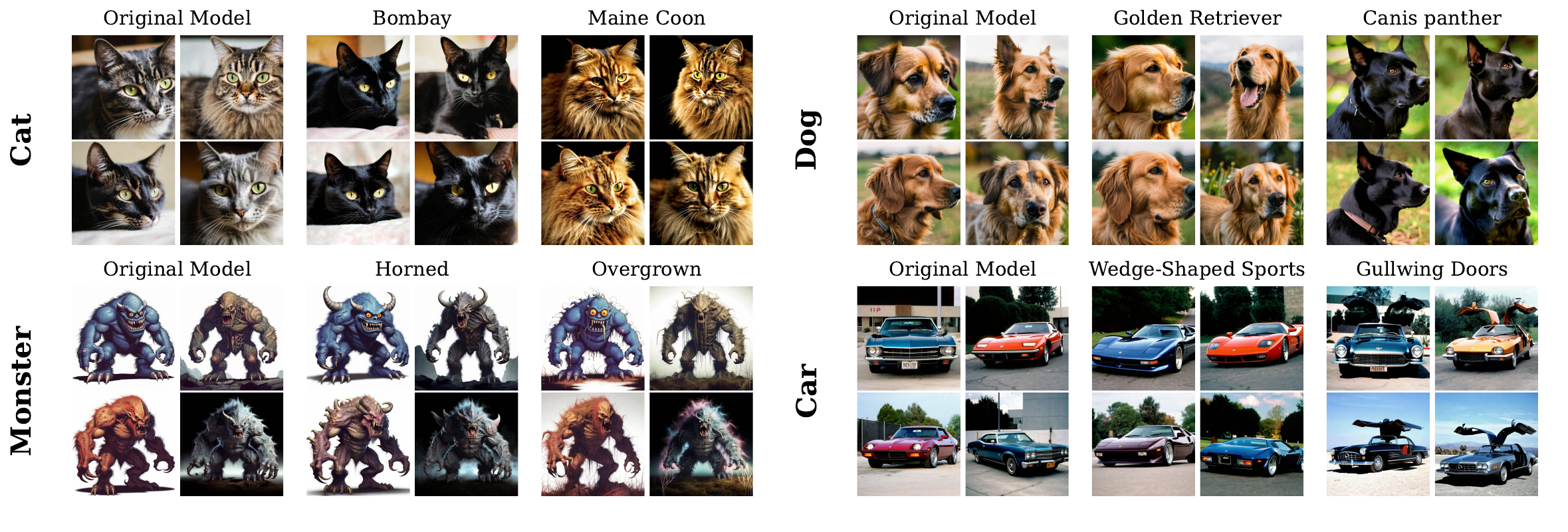}
    \caption{
    \ours{} decomposes the visual capabilities of the diffusion model. We show latent directions discovered in SDXL-DMD. 
    \vspace{-1em}
    }
    \label{fig:some_directions}
\end{figure}

Having established that the recovered subspace is valid, we turn to its practical use. Following~\citet{gandikota2025sliderspace}, we apply our method to a set of concepts, extract the interpretable latent directions, and use them to steer the generation process. As visualized in Fig.~\ref{fig:some_directions}, the discovered directions correspond to distinct, semantically coherent attributes, with further examples in
App.~\ref{app:details}.

\begin{figure*}[t] 
    \centering
    \includegraphics[width=\textwidth]{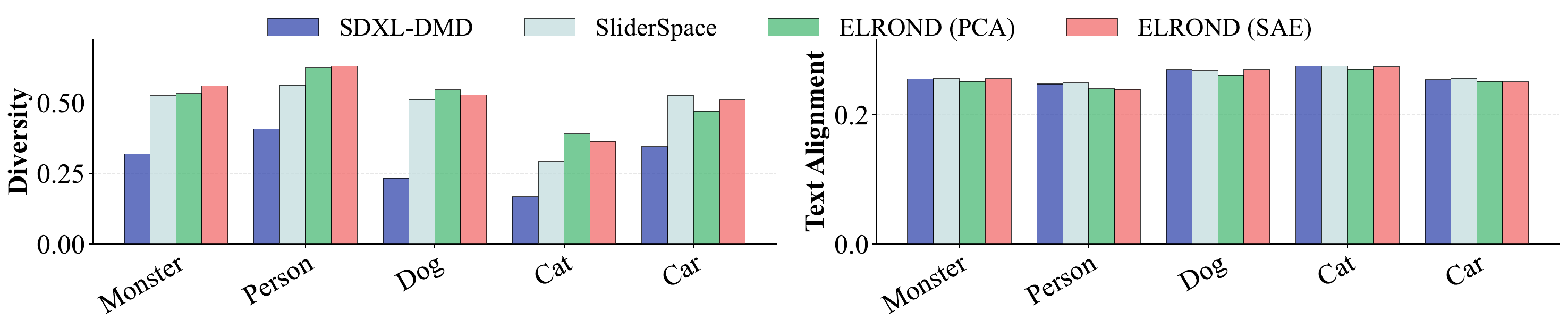}
    \vspace{-1em}
    \caption{
    \ours{} generates visually diverse variations of concepts (left) as measured by mean DreamSim distance (higher is better). Despite this, it maintains similar text-to-image alignment (right), as measured by CLIP score (higher is better). 
    }
    \label{fig:metrics_diversity}
\end{figure*}

A useful decomposition should not only identify attributes but also expose the model's full range of variation. To quantify this, we generate $10,000$ images per concept, modifying the concept token embedding of each by a linear combination of $3$ randomly selected latent directions. We measure intra-class diversity using DreamSim~\citep{fu2023dreamsim} and semantic consistency via CLIP Distance~\citep{hessel2021clipscore}, comparing against the base SDXL-DMD model and SliderSpace~\citep{gandikota2025sliderspace}. While all methods maintain high semantic consistency with the original prompt (Fig.~\ref{fig:metrics_diversity}, right), \ours{} yields a significantly higher diversity score than the base model across all tested concepts (Fig.~\ref{fig:metrics_diversity}, left) and outperforms SliderSpace. 

Since DreamSim measures perceptual distance without regard to what changed, a higher score could in principle reflect drift in background, composition, or style rather than variation of the concept itself. To rule this out, we sample $60$ image pairs per concept ($300$ per method) and ask Qwen2-VL~\citep{wang2024qwen2} to rate each pair on five independent axes - how much the concept's target attribute changed, and separately how much the composition, background, and style changed - together with an absolute artifact rating, and to make a single forced choice of the primary driver of the difference. As reported in Tab.~\ref{tab:drift_decomposition}, both variants of \ours{} rate the target attribute far above every off-target axis, artifacts remain low throughout, and the forced choice identifies the target attribute in over $96\%$ of pairs. The measured diversity gain therefore reflects genuine, concept-relevant variation rather than incidental drift.

A user study (Fig.~\ref{fig:elrond_vs}) confirms the advantage of our method: 40 human evaluators significantly preferred \ours{} in terms of diversity and creativity, while on image fidelity, our method's win rate falls within the confidence interval of being statistically equivalent to the unperturbed model, indicating no degradation in quality. 

We posit that this advantage stems from the source of the semantic signal. Unlike \ours, SliderSpace derives its axes from CLIP embeddings of generated images, and thus organizes the variations according to a model other than the one producing them. Consequently, the method is prone to missing modes of diversity that the external model struggles with, while detecting ones that the DM may be unable to produce (see App.~\ref{app:motivation_table} for discrepancies between the models).

\subsection{Diversity Enhancement and Mode Collapse Mitigation}
\label{subsec:diversity}
\vspace{-0.5em}

Having demonstrated that \ours{} is able to span the rich semantic manifold of learned concepts, we now apply it to the problem of mode collapse in distilled diffusion models~\citep{yin2024improved}, which manifests as repetitive outputs with reduced variation compared to their teachers. We discover latent directions for a set of concepts using both the teacher model (SDXL) and the distilled student (SDXL-DMD). To undo mode collapse, we sample images using SDXL-DMD while modifying the prompt embedding with a linear combination of 3 random directions. We evaluate three settings: using directions discovered within SDXL-DMD by PCA and SAE, and transferring directions discovered by SAE in the base SDXL model. We extend our evaluation to SDXL-Turbo in App.~\ref{app:sdxl_turbo}.

\begin{figure}[t]
    \centering
    \begin{minipage}[c]{0.45\textwidth}
        \centering
        \includegraphics[width=\linewidth]{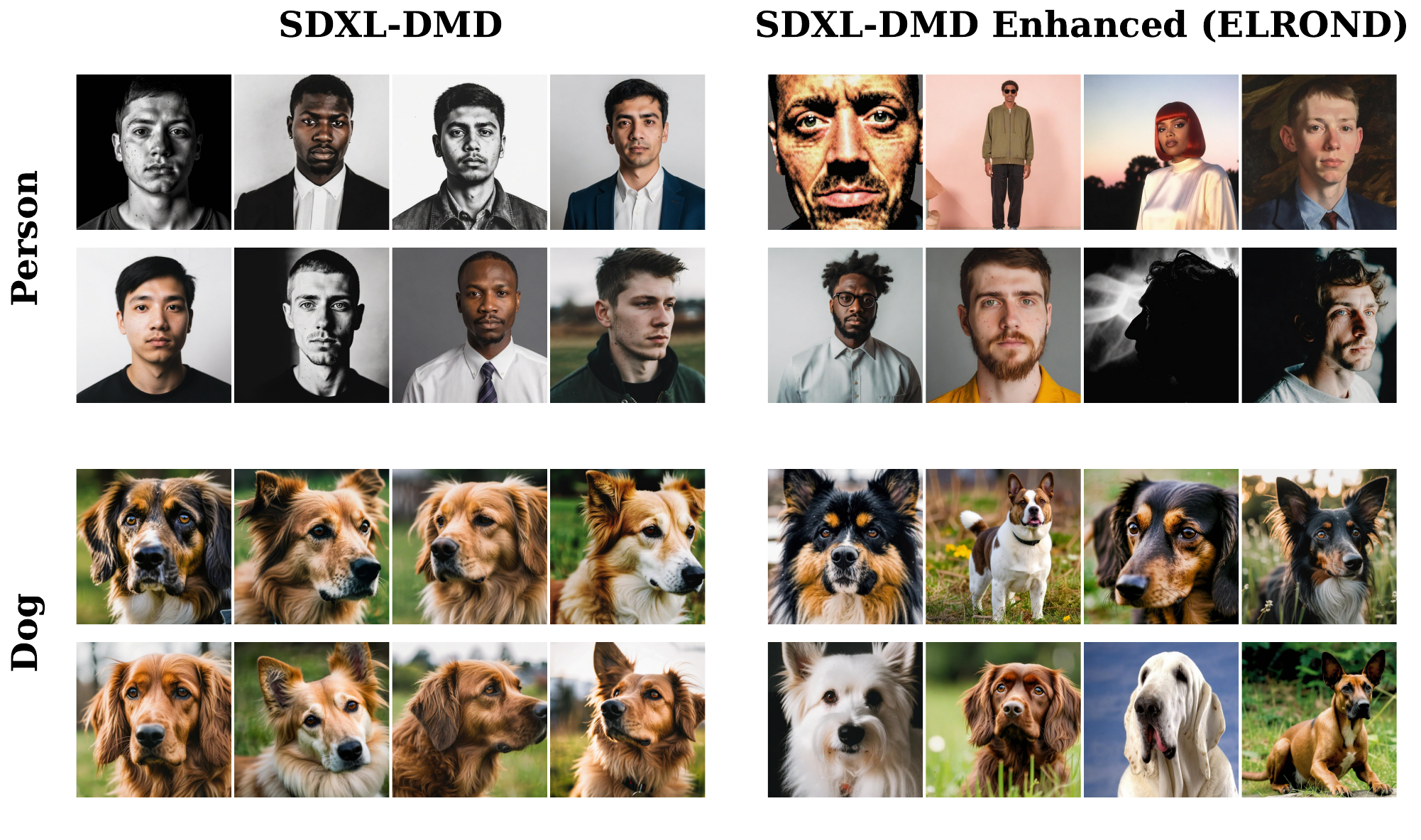}
        \vspace{-2.5em}
        \caption{\small Adding latent directions discovered by \ours{} (SAE) in the SDXL-DMD model, mitigates mode collapse in distilled model.}
        \label{fig:diversity_comparison}
    \end{minipage}\hfill
    \begin{minipage}[c]{0.53\textwidth}
        \centering
        \renewcommand\theadfont{\bfseries} 
        \renewcommand\theadgape{\Gape[4pt]} 
        \captionof{table}{Comparison of FID scores calculated between samples generated by SDXL and the distilled SDXL-DMD.}
        \label{tab:fid_results}
        \resizebox{\linewidth}{!}{%
        \begin{tabular}{lccccc}
            \toprule
            \thead{Concept} & \thead{ELROND \\ (SDXL)} & \thead{ELROND \\ (SDXL-DMD)} & \thead{ELROND PCA \\ (SDXL-DMD)} & \thead{Slider\\Space} & \thead{SDXL-\\DMD} \\ 
            \midrule
            Car     & \textbf{25.48} & \underline{26.35} & 26.57 & 31.80  & 37.55 \\
            Cat     & \textbf{19.72} & \underline{19.94} & 23.60 & 21.69  & 38.79 \\
            Dog     & \textbf{24.06} & \underline{25.13} & 32.95 & 38.25  & 50.25 \\
            Monster & \textbf{95.05} & \underline{99.63} & 107.46 & 136.89 & 103.95 \\
            Person  & \textbf{33.61} & \underline{38.90} & 47.35 & 41.52  & 86.94 \\
            \bottomrule
        \end{tabular}%
        }
    \end{minipage}
    \vspace{-1em}
\end{figure}

Qualitative results in Fig.~\ref{fig:diversity_comparison} show that \ours{} reverses the mode collapse. We quantify this in Tab.~\ref{tab:fid_results} by computing the FID between a reference distribution of $10,000$ images sampled from the teacher and the distributions generated by SDXL-DMD under various control methods. Our SAE variant using student directions achieves a lower FID than both the base distilled model and SliderSpace, while directions transferred from the teacher perform best overall. This indicates that the teacher retains a richer tangent space than the student, and that our method can transfer it without retraining.

The SAE-based variant also consistently outperforms the PCA-based one. We attribute this to the SAE's ability to capture the \textit{probability density} of the concept manifold, rather than just its geometric span. While PCA identifies a linear subspace where random sampling implies a uniform distribution over the manifold's extent, the SAE dictionary adapts to the non-uniform density of the semantic gradients. Formally, because the SAE minimizes the expected reconstruction error $\mathcal{L} = \mathbb{E}_{\mathbf{g} \sim p(\mathbf{g})} \|\mathbf{g} - \hat{\mathbf{g}}\|^2$ over the gradient distribution $p(\mathbf{g})$, it naturally allocates a higher density of feature vectors to regions of the manifold where $p(\mathbf{g})$ is high. Consequently, sampling random directions from the SAE dictionary effectively approximates sampling from the true underlying density $p(\mathbf{g})$, yielding a generated distribution that better matches the reference than the geometry-only approximation of PCA.

\section{Limitations}
\vspace{-0.5em}
\ours{}, while effective, is not without limitations. First, while our approach
avoids retraining, data collection is computationally extensive due to
backpropagating through the diffusion model. However, even under an equal
computational budget, \ours{} maintains competitive performance (App.~\ref{app:equal_compute}). We additionally evaluate the effect of
dataset size on the discovered subspace in App.~\ref{app:dataset_size}.
Second, our concept-level analysis relies on the common assumption that the
influence of a token is predominantly localized to the concept it represents.
Because token embeddings are contextual, perturbing one token can nevertheless
affect the visual realization of other tokens (App.~\ref{app:failure}).
Finally, some directions discovered by either SAE or PCA may not admit a clear
interpretation (App.~\ref{app:failure}).

\section{Conclusions}
\vspace{-0.5em}
We introduce \ours{}, a method for recovering the tangent space of the generative manifold of a diffusion model, by collecting gradients in its conditioning space. We validate the recovery against a known ground truth and behaviourally on large-scale models, and show that the recovered structure enables broader exploration of a model's capabilities than external proxies and mitigates mode collapse in distilled models. Ultimately, we offer a geometry-grounded account of a model's generative variety, derived from the model itself rather than imposed upon it.

\bibliography{iclr2027_conference}
\bibliographystyle{iclr2027_conference}

\appendix

\section{Synthetic validation details}
In this section, we provide details on the synthetic dataset validation performed in Sec.~\ref{subsec:toy}

\subsection{Dataset details}
\label{app:toy_dataset}

The synthetic dataset consists of $100{,}000$ RGB images of size $64 \times 64$,
each containing a single geometric shape: a triangle, a square, or a circle. Every
image is generated from a $9$-dimensional conditioning vector, whose structure is
given in Tab.~\ref{tab:conditioning_vector}. The first three coordinates one-hot
encode the shape, and the remaining six control its colour, size, and position.
Because we decide which of these attributes vary for each shape, the dimension of
the corresponding tangent space is known exactly.

\begin{table}[H]
\centering
\caption{Structure of the $9$-dimensional conditioning vector.}
\label{tab:conditioning_vector}
\small
\begin{tabular}{clc}
\toprule
Index & Coordinate & Range \\
\midrule
0 & $\mathrm{is\_triangle}$ & $\{0, 1\}$ \\
1 & $\mathrm{is\_square}$   & $\{0, 1\}$ \\
2 & $\mathrm{is\_circle}$   & $\{0, 1\}$ \\
3 & $r$                     & $[0, 1]$ \\
4 & $g$                     & $[0, 1]$ \\
5 & $b$                     & $[0, 1]$ \\
6 & $\mathrm{size}$         & $[0, 1]$ \\
7 & $x_{\mathrm{pos}}$      & $[0, 1]$ \\
8 & $y_{\mathrm{pos}}$      & $[0, 1]$ \\
\bottomrule
\end{tabular}
\end{table}

\paragraph{Attributes.}
The colour is given by three coordinates, one per channel, each mapped to an
intensity of at most $220$. To keep every shape clearly visible against the
background, we scale all three channels down proportionally whenever their sum
exceeds $580$, and replace combinations whose sum falls below $30$ with dark grey.
The size coordinate maps to a half-size between $6$ and $26$ pixels, and the two
position coordinates place the centre of the shape between pixels $16$ and $48$
along each axis. With these ranges, a shape may extend slightly past the edge of
the canvas, but most of its surface always remains visible.

\paragraph{From conditioning to distribution.}
An attribute coordinate does not set the rendered value directly. Instead, it
specifies the distribution from which that value is drawn, mirroring how a text
prompt specifies a distribution of images rather than a single one
(Sec.~\ref{subsec:toy}). Given a coordinate value $v$, the rendered value is
sampled uniformly from the window $[v - h(v),\, v + h(v)]$ with half-width
\begin{equation}
h(v) = \max\bigl(0,\; 0.4 - |v - 0.5|\bigr).
\end{equation}
At $v = 0.5$ the window is widest and covers $[0.1, 0.9]$. As $v$ moves toward
either end of its range, the window moves with it and narrows linearly, until it
closes at $v = 0.1$ and $v = 0.9$. From there on, the attribute is rendered
exactly at $v$. A conditioning with attributes near the middle of their ranges
therefore describes a broad distribution of images, while one with attributes
near the extremes describes an almost deterministic one
(Fig.~\ref{fig:samples_seeds}).

\paragraph{Varying attributes.}
Each shape varies in a different subset of attributes, listed in
Tab.~\ref{tab:gt}: the square varies only in colour, the triangle in colour and
size, and the circle in colour, size, and position. The remaining attributes are
held fixed for that shape. We treat each shape as a separate concept, and the
number of its varying attributes gives the ground-truth dimension of its tangent
space.

\begin{table}[H]
\centering
\caption{Varying attributes of each shape and the resulting ground-truth
dimension of its tangent space.}
\label{tab:gt}
\small
\begin{tabular}{llc}
\toprule
Shape & Varying attributes & Ground-truth dimension \\
\midrule
Square   & $r, g, b$ & 3 \\
Triangle & $r, g, b, \mathrm{size}$ & 4 \\
Circle   & $r, g, b, \mathrm{size}, x_{\mathrm{pos}}, y_{\mathrm{pos}}$ & 6 \\
\bottomrule
\end{tabular}
\end{table}

\subsection{Model details}
\label{app:toy_model}

\paragraph{Architecture.}
The diffusion model is a UNet operating directly in pixel space on $64 \times 64$
RGB images. It has four resolution stages with $32$, $64$, $128$, and $128$
channels and two layers per block. The two highest resolutions ($64 \times 64$
and $32 \times 32$) use standard convolutional blocks, while the lower
resolutions ($16 \times 16$, $8 \times 8$, and $4 \times 4$) use cross-attention
blocks with a head dimension of $8$. The conditioning vector is mapped into the
model by the linear encoder $E$, which projects its $9$ coordinates to a
$64$-dimensional embedding matching the cross-attention dimension of the UNet.
This embedding space plays the role of the text embedding space of large-scale
models, and it is the space in which we collect gradients.

\begin{figure}[t]
    \centering
    \includegraphics[width=0.75\columnwidth]{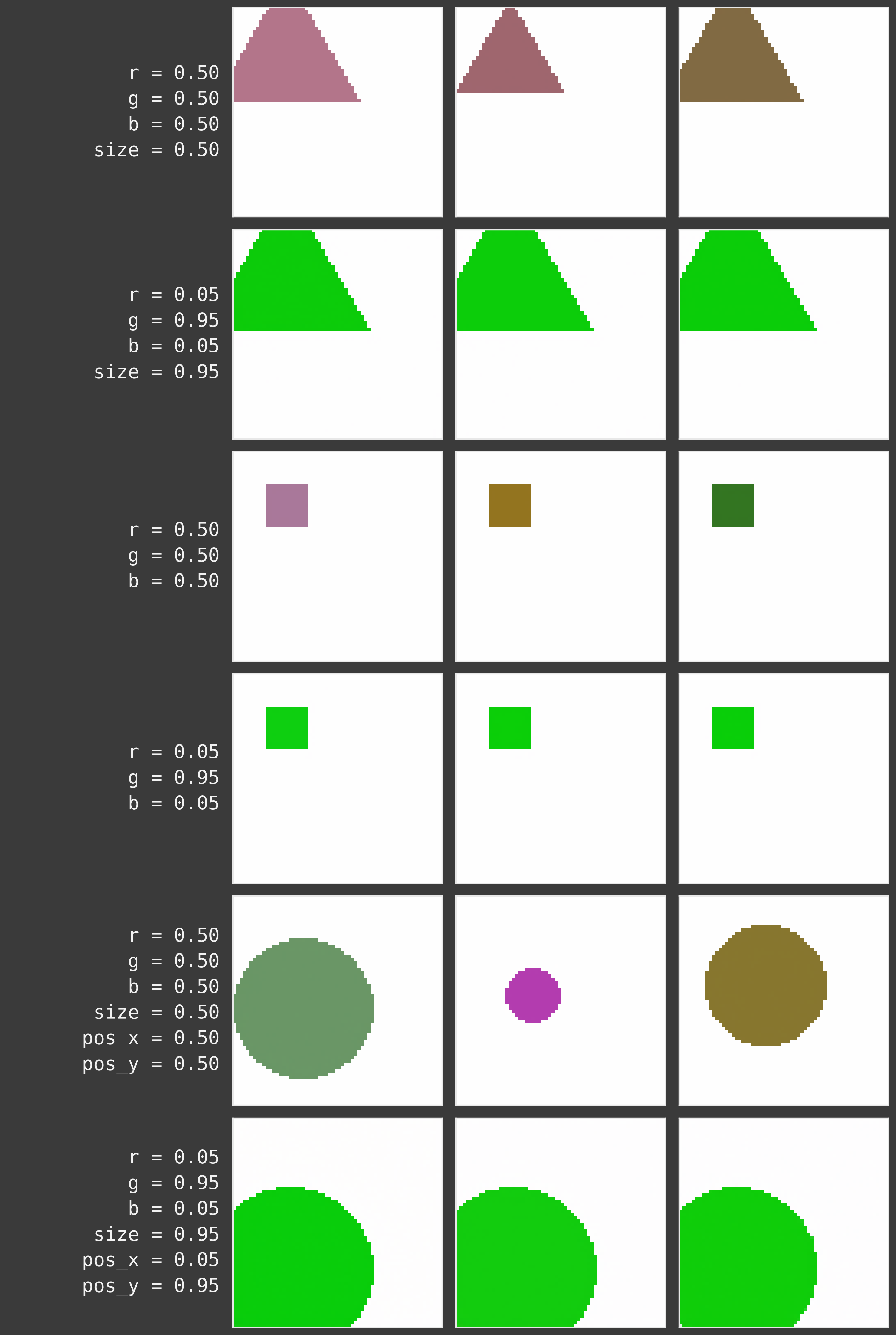}
    \caption{Samples from the trained diffusion model. Each row uses a single
    conditioning vector, shown on the left, and differs only in the noise seed.
    When all varying attributes are set to $0.5$, the sampling window is at its
    widest and the samples differ visibly. When they are set near the ends of
    their ranges, the window is closed and the samples are nearly identical. The
    number of listed attributes matches the number of varying attributes of each
    shape: three for the square, four for the triangle, and six for the circle.}
    \label{fig:samples_seeds}
\end{figure}

\paragraph{Training.}
We use a linear noise schedule with $1{,}000$ timesteps and train the model to
predict the added noise. Training runs for $200$ epochs with a batch size of
$64$, using AdamW with a learning rate of $10^{-4}$, $500$ warmup steps, and
gradient clipping at a norm of $1.0$. We additionally maintain an exponential
moving average of the model parameters with a decay of $0.9999$, starting after
$100$ training steps. Samples from the trained model are shown in
Fig.~\ref{fig:samples_seeds}.

\paragraph{Gradient collection and dimension estimation.}
For each shape, we generate $5{,}000$ images and use them to collect $10{,}000$
gradients at each of $20$ uniformly spaced timesteps. To estimate the dimension
of the recovered subspace, we train VAEase with a hidden width and latent
dimension of $64$, equal to the dimension of the gradients. Training runs for
$1{,}000$ epochs using Adam with a learning rate of $10^{-3}$ and a batch size of
$1{,}024$, with $5\%$ of the gradients held out for validation.

\subsection{Gradients reflect attribute differences}
\label{app:toy_gradient_correlation}

We verify that individual gradients carry the
information we expect. A gradient computed from a pair of images should reflect
the attributes that actually differ between them, rather than arbitrary
variation. We test this on the circle, the only shape for which all six
attributes vary.

We run $100$ independent trials. In each, we draw two circles, $A$ and $B$, with
independent attributes, and render both directly from their attribute values
using the dataset generator rather than sampling them from the model. We noise
the image of $A$ to level $t$ and use a single model call to estimate the clean
image. We then compute the pixel MSE between this estimate and the image of $B$,
and take its gradient with respect to the output of $E$. The model is always
conditioned on the same neutral conditioning, with every continuous attribute set
to $0.5$, rather than on the attributes of $A$. This way, all gradients are
computed at the same point in conditioning space.

For each attribute, we measure the alignment of the negated gradient with that
attribute's ground-truth direction, the image under $E$ of its coordinate
direction, via their dot product. We average this alignment over $200$ evenly
spaced timesteps. Since the gradient is taken at the neutral conditioning, the
expected change in each attribute is the difference between the value of $B$ and
$0.5$. We therefore correlate the averaged alignment with this difference across
the $100$ trials.

\begin{table}[h]
\centering
\caption{Correlation, across $100$ trials, between the alignment of the gradient
with each ground-truth direction and the difference between the attribute value
of circle $B$ and $0.5$. Alignments are averaged over $200$ evenly spaced
timesteps before computing the correlation.}
\label{tab:gradient_correlation}
\begin{tabular}{lc}
\toprule
Attribute & Correlation \\
\midrule
Red          & 0.509 \\
Green        & 0.374 \\
Blue         & 0.576 \\
Size         & 0.678 \\
$x$ position & 0.737 \\
$y$ position & 0.719 \\
\bottomrule
\end{tabular}
\end{table}

As shown in Tab.~\ref{tab:gradient_correlation}, all six attributes correlate
positively, confirming that the gradients track the attributes that differ
between the images. The correlations are moderate rather than close to one. We
attribute this mainly to averaging over the entire sampling trajectory: attributes
are not equally determined at all noise levels, so timesteps at which an
attribute is barely affected by the conditioning dilute its signal.

\subsection{Dimensionality discrepancy for the circle}
\label{app:toy_discussion}

For the circle, VAEase estimates an $8$-dimensional subspace, whereas the dataset
defines only $6$ free generative factors. We argue that this discrepancy reflects
the trained model rather than a failure of \ours{}. Our ground truth rests on the
assumption, stated in Sec.~\ref{subsec:toy}, that the model has learned the data
distribution faithfully. Under that assumption, the only conditioning directions
that alter its output are the images under $E$ of the varying attributes.
\ours{}, however, recovers $\mathcal{T}_c = \ker(d\Phi_c)^\perp$, which is defined
by the model and not by the dataset. If the model has additionally learned to
respond to conditioning directions that correspond to no generative factor, those
directions belong to $\mathcal{T}_c$, and a correct recovery should include them.
Nothing in training rules this out. The model only ever observes conditionings
from a $9$-dimensional subspace of its $64$-dimensional input space, and its
response to other directions is left unconstrained.

The precision and recall reported in Tab.~\ref{tab:toy_results} already point in
this direction. Both quantities are normalised traces of $P\hat{P}$, so they
satisfy $\dim\hat{\mathcal{T}}_c \cdot \text{precision} = \dim\mathcal{T}_c \cdot
\text{recall}$. The circle's recall of $0.968$ shows that the ground-truth
subspace is almost entirely contained in the recovered one. Its precision of
$0.726$ equals $0.968 \cdot 6/8$. The drop in precision is therefore fully
accounted for by the two surplus dimensions, not by recovered directions that
deviate from the ground truth. What remains to establish is whether these two
dimensions are genuine structure of the model or estimation noise.

To test whether the model genuinely responds to these two dimensions, we first
isolate them. Given matrices $V \in \mathbb{R}^{8 \times 64}$ and
$A \in \mathbb{R}^{6 \times 64}$ whose orthonormal rows span the recovered and
ground-truth subspaces, we compute
\begin{equation}
R = V - V A^\top A,
\end{equation}
which removes from the recovered subspace everything contained in the ground
truth. The two leading right singular vectors of $R$ form an orthonormal basis of
the part of the recovered subspace orthogonal to the ground truth. We refer to
them as the additional directions.

We then intervene along each direction and measure how much the generated image
changes. For each of $100$ conditioning vectors of the circle, we shift the
conditioning along a given direction with strength $\lambda$. We then compare the
resulting image with the one generated from the unshifted conditioning under the
same seed, using the mean squared pixel difference. As a reference, we repeat the
procedure for two random directions of equal norm, orthogonal to the full
$8$-dimensional recovered subspace. According to our method, the model should be
insensitive to such directions. If the additional directions were an artifact of
the estimation, they should behave in the same way.

\begin{table}[h]
\centering
\caption{Mean squared pixel difference between images generated from shifted and
unshifted conditionings of the circle, for increasing intervention strength.
Results are averaged over $100$ conditioning vectors with matched seeds and over
the two directions in each group. The last column gives the ratio between the
additional and random directions.}
\label{tab:residual_mse}
\begin{tabular}{lccc}
\toprule
Strength & Additional & Random & Ratio \\
\midrule
0.1 & 0.0035 & 0.00058 & 6.1 \\
0.2 & 0.0071 & 0.00094 & 7.6 \\
0.3 & 0.0109 & 0.00149 & 7.3 \\
0.4 & 0.0144 & 0.00184 & 7.8 \\
0.5 & 0.0177 & 0.00224 & 7.9 \\
0.6 & 0.0218 & 0.00249 & 8.8 \\
\bottomrule
\end{tabular}
\end{table}

As shown in Tab.~\ref{tab:residual_mse}, the additional directions change the
generated image $6$ to $9$ times more than random directions of the same norm, at
every strength we tested. The model therefore does respond to them, and by
definition they belong to the tangent space of the trained model. The recovered
$8$-dimensional subspace is thus an accurate description of the model; the
discrepancy with the ground truth stems from the model not having learned the
dataset perfectly.

\section{Linearity in Semantic Latent Spaces}
\label{app:motivation_table}
In this section, we show that the organization of visual variations induced by an external vision model differs significantly from that of a diffusion model. 

\begin{figure}[h]
        \centering
        \begin{subfigure}[b]{0.48\linewidth}
            \centering
            \includegraphics[width=\textwidth]{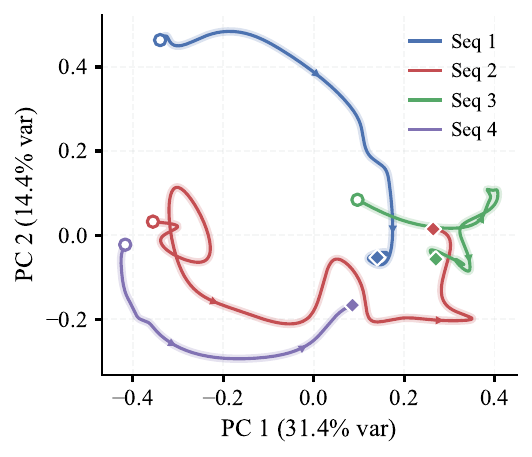} 
            \label{fig:motivation_trajectories}
        \end{subfigure}
        \begin{subfigure}[b]{0.48\linewidth}
            \centering
            \includegraphics[width=\textwidth]{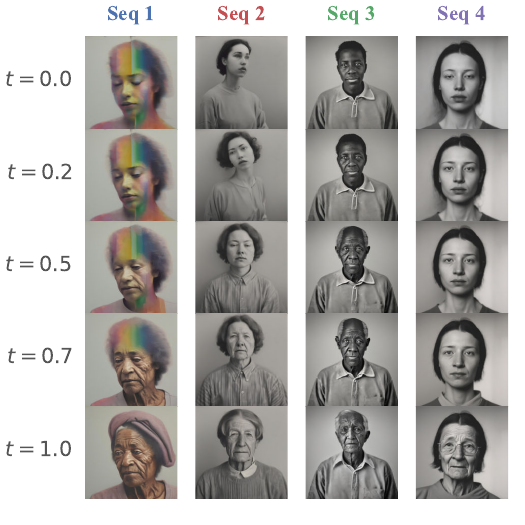}
            \label{fig:motivation_images}
        \end{subfigure}
    
        \caption{
        Semantic transitions in the diffusion model outputs resulting from linear interpolation (right) do not correspond to a linear direction within CLIP embedding space (left) (first 2 PCA dimensions). 
        }
        \label{fig:motivation}
\end{figure}

That these two organizations disagree can be shown directly. Following \cite{baumann2025continuous}, we traverse the conditioning space of SDXL along a straight path between the embeddings of $c_{start} = \text{``Person''}$ and $c_{end} = \text{``Old person''}$, generating images conditioned on $c_t = (1-t)c_{start} + t c_{end}$ with the noise held fixed, and project the resulting sequence into the CLIP visual embedding space. As shown in Fig.~\ref{fig:motivation}, a displacement that is uniform for the diffusion model and perceptually smooth in the images is strongly curved under CLIP. Steps of equal size for one model are therefore of unequal size for the other, and axes that are orthogonal or evenly spaced under an external proxy carry no such guarantee with respect to the geometry of the diffusion model.

To quantify this, we generated $N=700$ trajectories between distinct semantic concepts. We projected these trajectories into the CLIP visual embedding space and computed four metrics to quantify their deviation from linearity:
\begin{enumerate}
    \item \textbf{Straightness Index:} The ratio of the Euclidean distance between the start and end points to the cumulative length of the trajectory. A value of $1.0$ indicates a perfectly straight line.
    \item \textbf{Average Cosine Similarity:} The mean cosine similarity between consecutive step vectors along the path. 
    \item \textbf{Variance Ratio (PC1):} The proportion of variance explained by the first principal component of the trajectory points.
    \item \textbf{RMSE (Linear Fit):} The root mean square error of the trajectory points relative to a linear interpolation between the start and end states.
\end{enumerate}

The results, presented in Table \ref{tab:linearity_metrics}, demonstrate that simple linear interpolation in the text conditioning space yields highly non-linear paths in the visual embedding space, necessitating the gradient-based approach proposed in our method.

\begin{table}[h]
    \centering
    \caption{\textbf{Quantitative Analysis of Trajectory Linearity.} 
    Metrics are averaged over $N=700$ trajectories. 
    The ``Ideal'' column indicates values for a perfectly linear path. 
    The observed negative cosine similarity and low straightness index confirm the highly non-linear nature of the latent evolution.}
    \label{tab:linearity_metrics}
    
        \begin{tabular}{l c c}
            \toprule
            \textbf{Metric} & \textbf{Observed Value} & \textbf{Ideal (Linear)} \\
            \midrule
            Avg. Cosine Similarity & $-0.4696 \pm 0.0590$ & $1.0$ \\
            Straightness Index     & $0.1696 \pm 0.0180$  & $1.0$ \\
            Variance Ratio (PC1)   & $0.3353 \pm 0.0472$  & $1.0$ \\
            RMSE (Linear Fit)      & $0.4271 \pm 0.0254$  & $0.0$ \\
            \bottomrule
        \end{tabular}
\end{table}

\section{SAE Dictionary Size}
\label{app:sae_ablation}

The expansion factor of the SAE determines the size of the learned dictionary and therefore the number of directions the decomposition can express. To select it, we fix the concept (\textit{Monster}) and the sparsity $k=32$, train SAEs at expansion factors $\{1, 2, 4, 8, 16, 32\}$ on the same $50{,}000$ gradients, and count the latents that are selected in the top-$k$ activation of at least one
input.

\begin{table}[h]
\centering
\caption{Active latents as a function of dictionary size, for a fixed sparsity
$k=32$ on the \textit{Monster} concept.}
\label{tab:sae_ablation}
\small
\begin{tabular}{cccc}
\toprule
Expansion factor & Dictionary size & Active latents & \% active \\
\midrule
1  & 2,048  & 1,497 & 73.1\% \\
2  & 4,096  & 2,228 & 54.4\% \\
4  & 8,192  & 3,151 & 38.5\% \\
8  & 16,384 & 4,488 & 27.4\% \\
16 & 32,768 & 6,523 & 19.9\% \\
32 & 65,536 & 9,228 & 14.1\% \\
\bottomrule
\end{tabular}
\end{table}

As reported in \cref{tab:sae_ablation}, the number of active latents grows far more slowly than the dictionary itself, increasing roughly sixfold over a thirty-twofold range of capacity, with the fraction of used latents falling from $73\%$ to $14\%$. Additional capacity is therefore absorbed largely as unused dictionary entries rather than converted into new directions, and we adopt an expansion factor of $1$ throughout as the most efficient point on this curve.

\section{Intervention Strength}
\label{app:intervention_str}

Throughout the main experiments we steer by adding a linear combination of three randomly selected directions, scaled by an intervention strength $\lambda = 40$ for all concepts except \textit{Person}, for which $\lambda = 60$ performed better empirically. To characterize this choice, we sweep $\lambda$ from $1$ to $100$ on the \textit{Monster} concept for both PCA and SAE directions, measuring intra-pool DreamSim as a measure of diversity and CLIP score against the prompt as a measure of alignment, with $2{,}000$ images per point.

\begin{table}[h]
\centering
\caption{Diversity and prompt alignment as a function of intervention strength
$\lambda$ on the \textit{Monster} concept.}
\label{tab:intervention_strength}
\small
\begin{tabular}{lcccccccc}
\toprule
$\lambda$ & 1 & 5 & 10 & 20 & 30 & 50 & 70 & 100 \\
\midrule
PCA, DreamSim $\uparrow$   & 0.318 & 0.333 & 0.363 & 0.445 & 0.529 & 0.635 & 0.690 & 0.747 \\
PCA, CLIP score $\uparrow$ & 0.265 & 0.266 & 0.265 & 0.260 & 0.253 & 0.243 & 0.237 & 0.231 \\
\midrule
SAE, DreamSim $\uparrow$   & 0.321 & 0.325 & 0.349 & 0.406 & 0.462 & 0.572 & 0.648 & 0.720 \\
SAE, CLIP score $\uparrow$ & 0.266 & 0.265 & 0.265 & 0.262 & 0.258 & 0.250 & 0.243 & 0.235 \\
\bottomrule
\end{tabular}
\end{table}

Both metrics vary smoothly and monotonically across the entire range for both decompositions, exhibiting no threshold or discontinuity at which the intervention begins to break down. Alignment is essentially unaffected below $\lambda = 10$, where diversity also grows slowly, and declines steadily thereafter as diversity increases. The values used in the main experiments sit past this inert region while alignment remains close to its maximum. We do not claim this operating point is optimal, only that it is a reasonable choice on a smooth trade-off admitting no sharper criterion.

\clearpage
\section{Influence of the dataset size on the resulting space}

\label{app:dataset_size}
In this section, we analyze the stability of the discovered semantic subspace relative to the number of gradients, $N$, used for decomposition. Our objective is to determine a sufficient dataset size that captures the concept's manifold without excessive computational cost.

To evaluate this, we compute the Principal Component Analysis (PCA) multiple times using varying subset sizes and compare the resulting subspaces against a reference space constructed from $50,000$ gradients.

\begin{figure}[h!] 
    \centering
    \includegraphics[width=\linewidth]{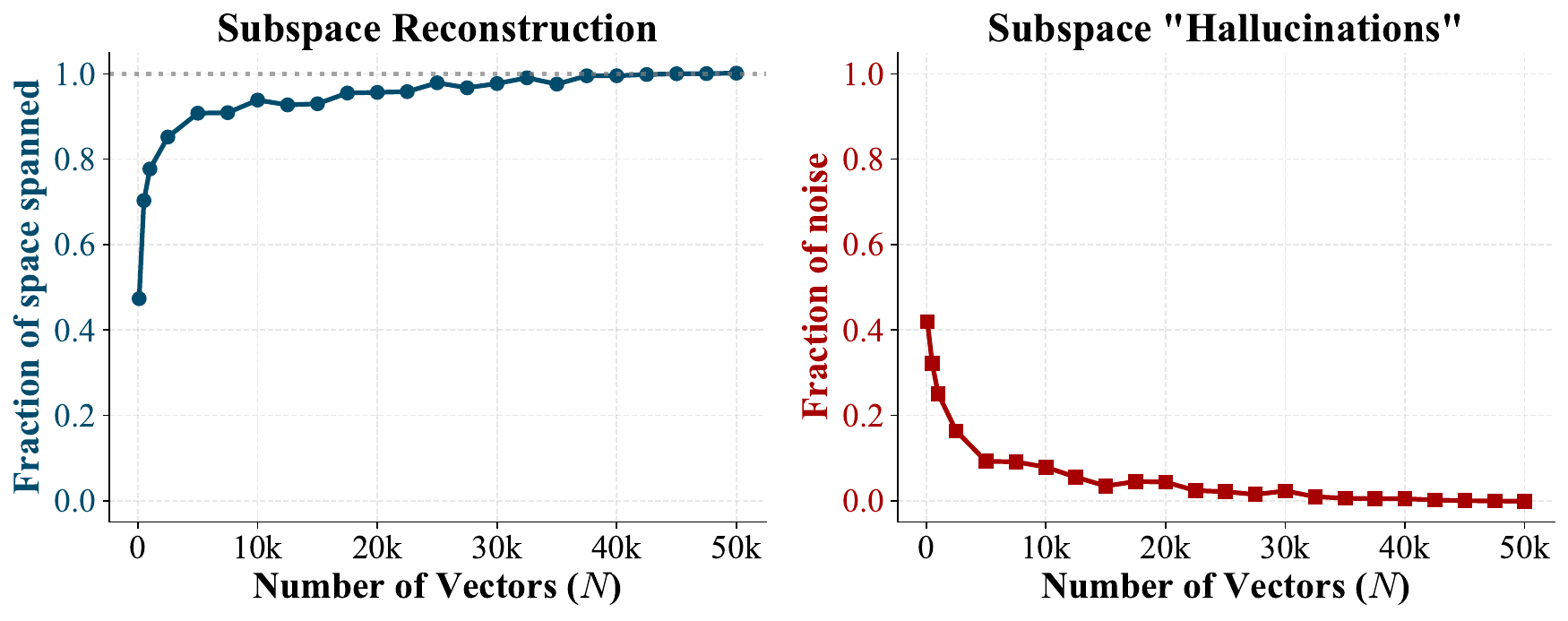}
    \vspace{-1em}
    \caption{Influence of number of gradients on resulting subspace  
    }
    \label{fig:num_grad_analysis}
\end{figure}

As illustrated in \cref{fig:num_grad_analysis}, the discovered subspace stabilizes rapidly as the number of vectors increases. In \cref{fig:num_grad_analysis} (left), we measure the fraction of subspace spanned by the smaller subset compared to the reference space. The curve shows a steep initial ascent, indicating that even a small number of gradients can capture the primary semantic directions.
On the other hand in \cref{fig:num_grad_analysis} (right) we quantify the fraction of noise, defined as directions present in the smaller subset but absent in the reference, to ensure the discovered axes are semantically valid. This noise fraction decreases significantly as $N$ approaches $30,000$.

By comparing the trade-off between reconstruction accuracy and the reduction of ''hallucinated'' directions, we identify $30,000$ gradients as the optimal sweet spot for our experiments. This volume provides a robust representation of the concept manifold while maintaining computational efficiency.

Additionally, to verify that the discovered semantic subspace is invariant to the initial noise sampling, we evaluate the stability of \ours{} across different random seeds. We sample two independent datasets of $5,000$ images initialized with distinct random noise, calculate $25,000$ semantic gradients for each set, and extract their respective subspaces via PCA. Averaged over $10$ such independent pairs, the fraction of the subspace spanned by one set relative to the other remains highly robust ($0.930 \pm 0.01$), accompanied by a negligible fraction of noise ($0.069 \pm 0.01$). These results confirm that \ours{} successfully captures the intrinsic semantic manifold of a given concept, fundamentally independent of the specific noise initialization used during the dataset collection phase.

\clearpage
\section{Diffusion timestep selection}
\label{app:timesteps_selection}

\begin{figure}[h] 
    \centering
    \includegraphics[width=.7\linewidth]{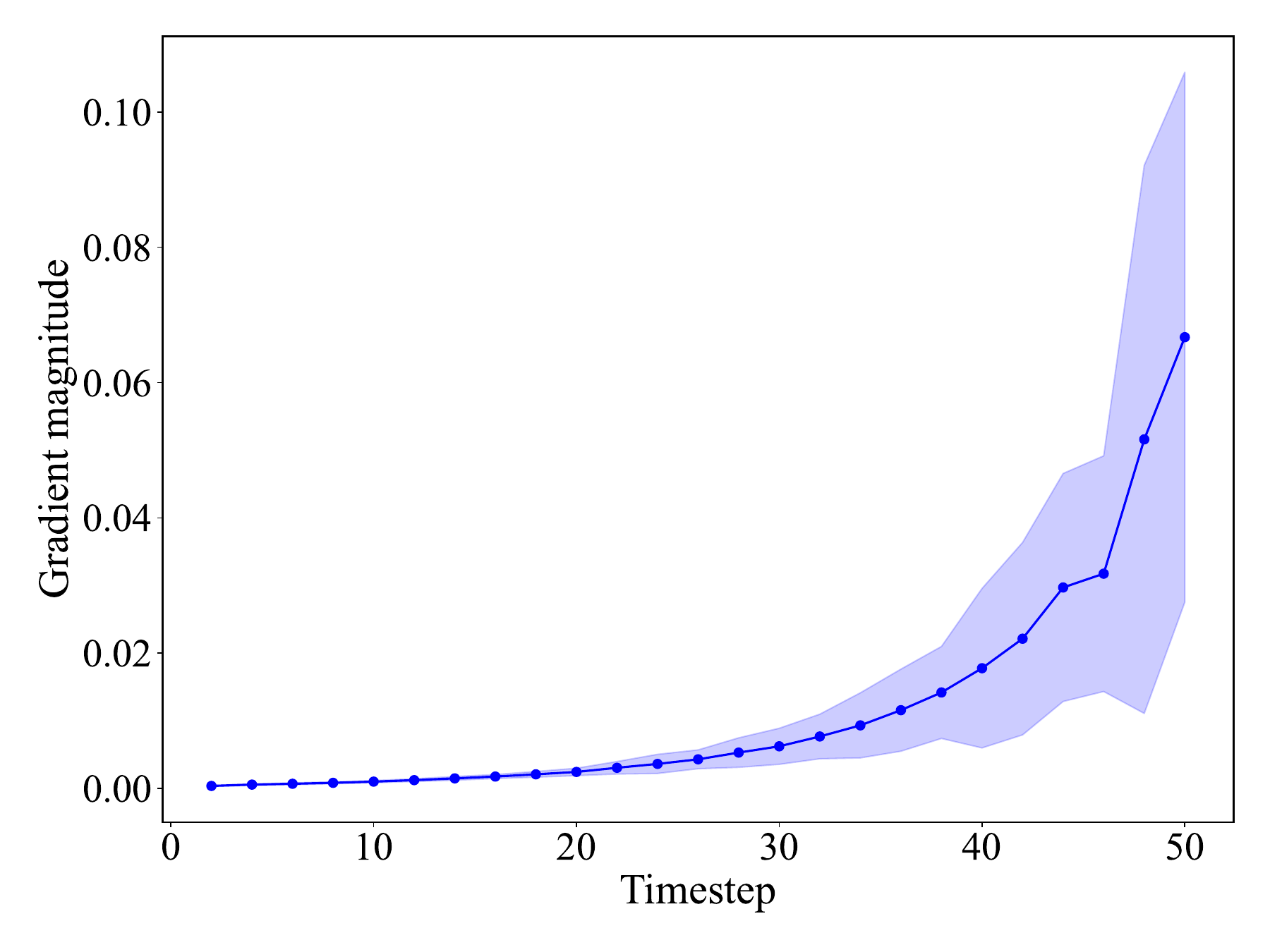}
    \caption{\textbf{Gradient Magnitude across Diffusion Timesteps.} The evolution of semantic signal strength relative to the denoising schedule. The increase in gradient magnitude at higher timesteps suggests that the model’s receptivity to text-based conditioning is most significant during the early stages of generation, where the high noise floor requires the conditioning signal to define the core semantic layout.
    }
    \label{fig:grad_norm}
\end{figure}

One of the hyperparameters affecting the way our method works is the selection of the diffusion timestep $t$ at which semantic gradients are collected. In this section, we analyze how the noise level during backpropagation influences the richness and magnitude of the discovered semantic directions. Figure~\ref{fig:grad_norm} illustrates the relationship between the diffusion timestep and the magnitude of the resulting gradient backpropagated to the specific token. We observe that the gradient magnitude generally increases with $t$, peaking at the highest noise levels. As shown in Table \ref{tab:results_timesteps}, we quantitatively evaluated the semantic alignment of these gradients using DreamSim, DINO, and CLIP similarity metrics.

Our analysis indicates that gradients collected at high noise levels ($t \approx T$) produce the strongest CLIP and DINO similarity scores. This is in line with recent work by~\cite{liu2024faster}, which suggests that during the initial stages of the diffusion process, the model is most reliant on text conditioning to establish the global structure of the image. While the DreamSim distance remains relatively consistent across various timesteps, there is a marginal improvement as $t$ increases, which points toward more precise semantic transitions. Based on these observations, we identified the early denoising stages as the optimal spot for capturing the most semantically rich and steerable directions within the embedding space

\begin{table}[h]
    \centering
    \begin{tabular}{lccc}
    \toprule
    Timestep & DreamSim ($\downarrow$) & DINO ($\uparrow$) & CLIP ($\uparrow$) \\
    \midrule
    36  & 0.415 & 0.485 & 0.781 \\
    40  & 0.415 & 0.488 & 0.782 \\
    43  & 0.415 & 0.486 & 0.783 \\
    46  & 0.407 & 0.491 & 0.785 \\
    50  & \textbf{0.404} & \textbf{0.502} & \textbf{0.791} \\
    \bottomrule
    \end{tabular}
    \caption{Average (across 500 pairs) DreamSim distance, DINO similarity, and CLIP similarity between image changed by a gradient from the corresponding timestep and the image towards which the gradient is pointing. Arrows indicate the direction of better performance.}
    \label{tab:results_timesteps}
\end{table}

\clearpage


\section{Generalization to Transformer-based Models}
\label{app:flux}
\begin{figure}[h] 
    \centering
    \includegraphics[width=\linewidth]{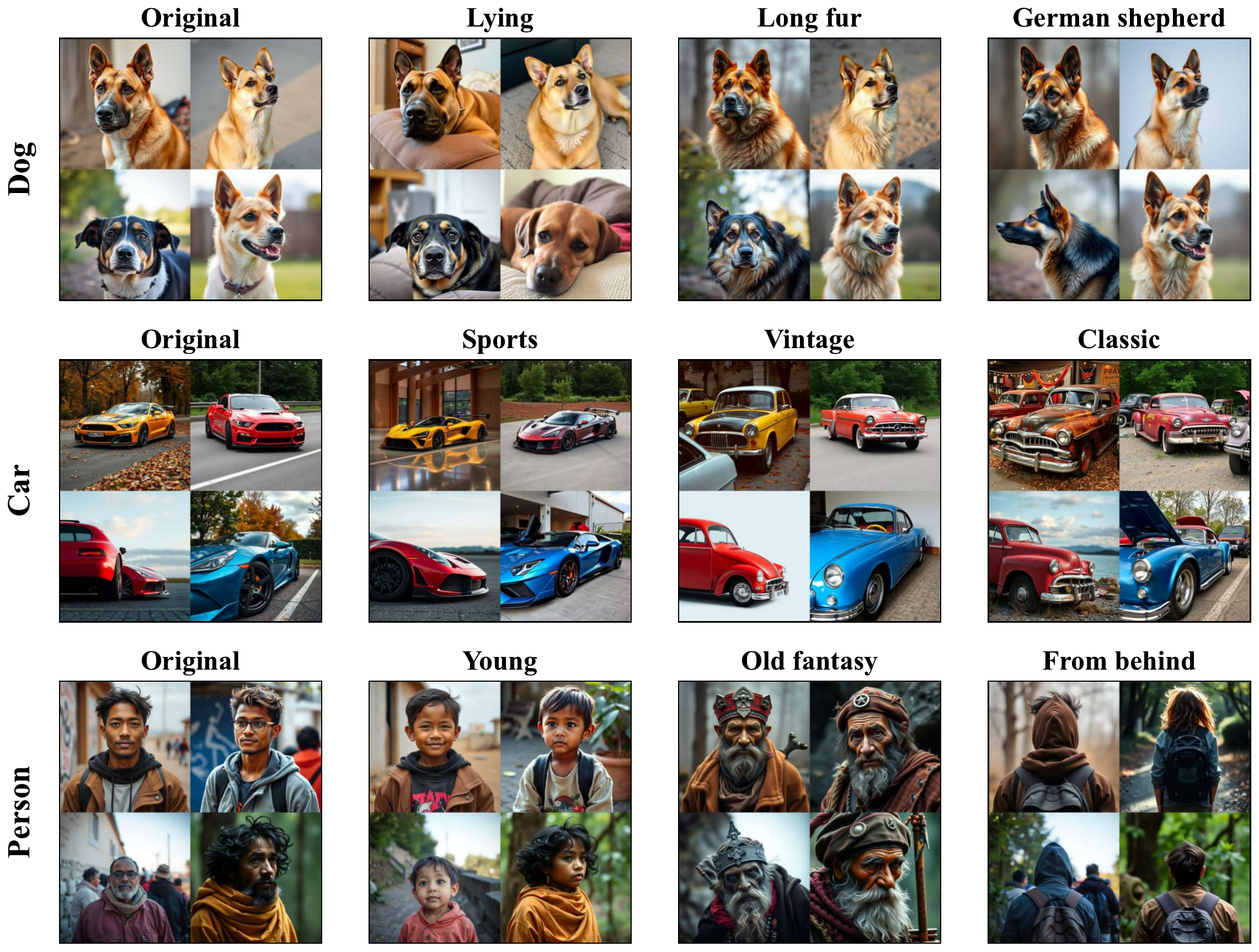}
    \vspace{-1em}
    \caption{\textbf{Generalization to Transformer-based Architectures.} \ours{} effectively extracts semantically coherent latent directions from Flux Schnell, a transformer-based diffusion model. The discovered directions for Dog, Car, and Person show high levels of interpretability. }
    \label{fig:flux_examples}
\end{figure}

To evaluate the architectural agnosticism of \ours{}, we extended our experiments to Flux Schnell, a state-of-the-art transformer-based diffusion model. As illustrated in Figure \ref{fig:flux_examples}, our method successfully identifies semantically rich and steerable directions within the conditioning space of the transformer architecture. The discovered directions for the tested concepts show high levels of semantic coherence, including different styles of cars, textures/breeds of dogs, or demographic and environmental attributes. Interestingly, our method was also able to discover composition-related features like the pose of a dog (lying) or a person (from behind).

\clearpage
\section{Analysis of Failure Cases and Limitations}

\begin{figure}[h] 
    \centering
    \includegraphics[width=.8\linewidth]{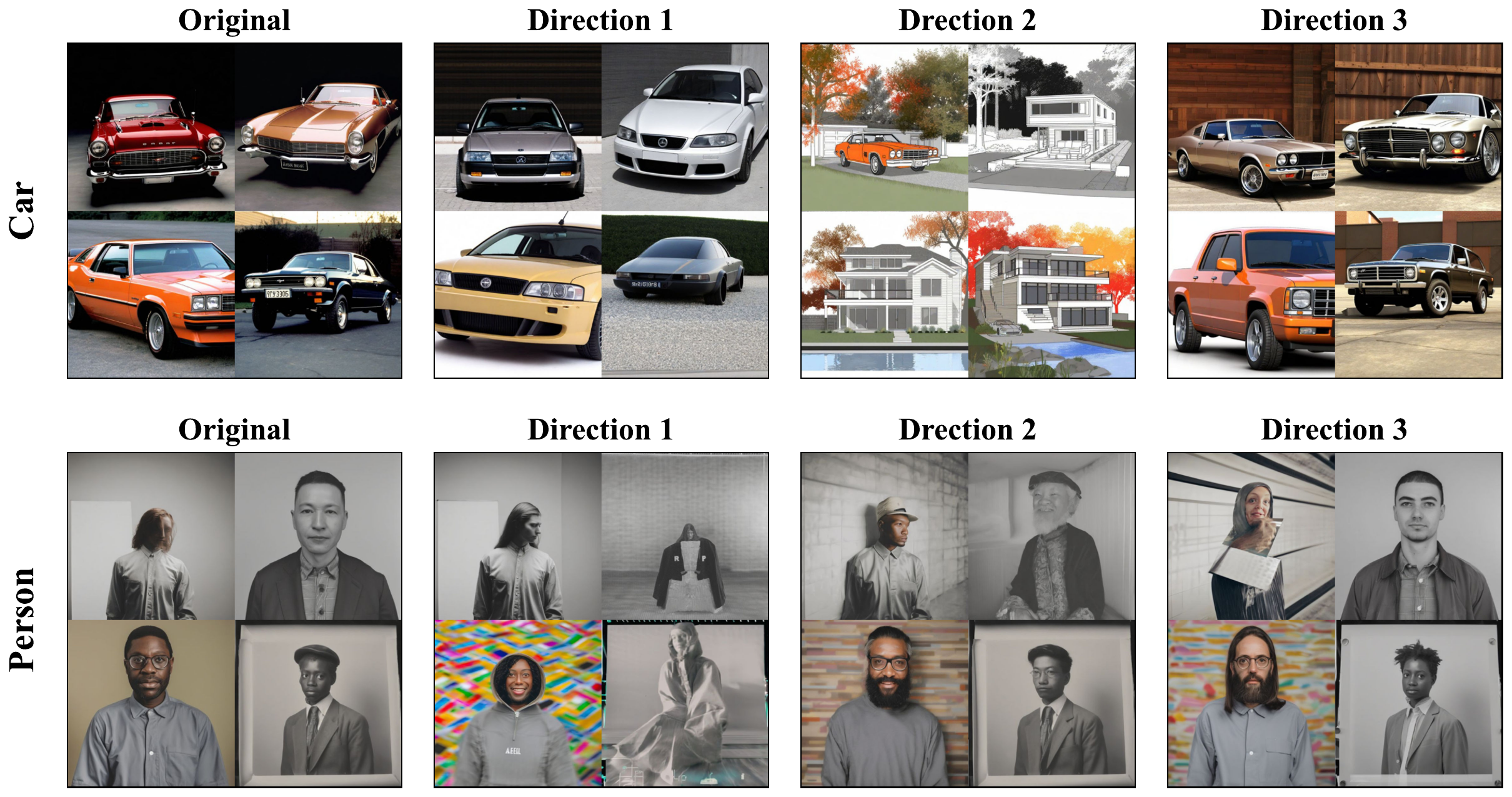}
    \caption{Failure cases for sdxl-dmd.
    }
    \label{fig:failure_case}
\end{figure}

As discussed in the limitations section, the decomposition of the gradient dataset $\mathcal{G}$ can occasionally yield directions that lack clear human interpretability. As visualized in Figure \ref{fig:failure_case}, these failure modes typically fall into two categories:
\begin{itemize}
    \item Perceptually Inert Directions: Modulations that, despite a non-zero norm, fail to produce a significant semantic shift in the generated output.
    \item Unstructured Artifacts: Interventions that trigger "hallucinated" textures or chaotic pixel-level variations rather than coherent attribute transformations.
\end{itemize}

\begin{figure}[h!] 
    \centering
    \begin{subfigure}[b]{0.49\textwidth}
        \centering
        \includegraphics[width=.8\textwidth]{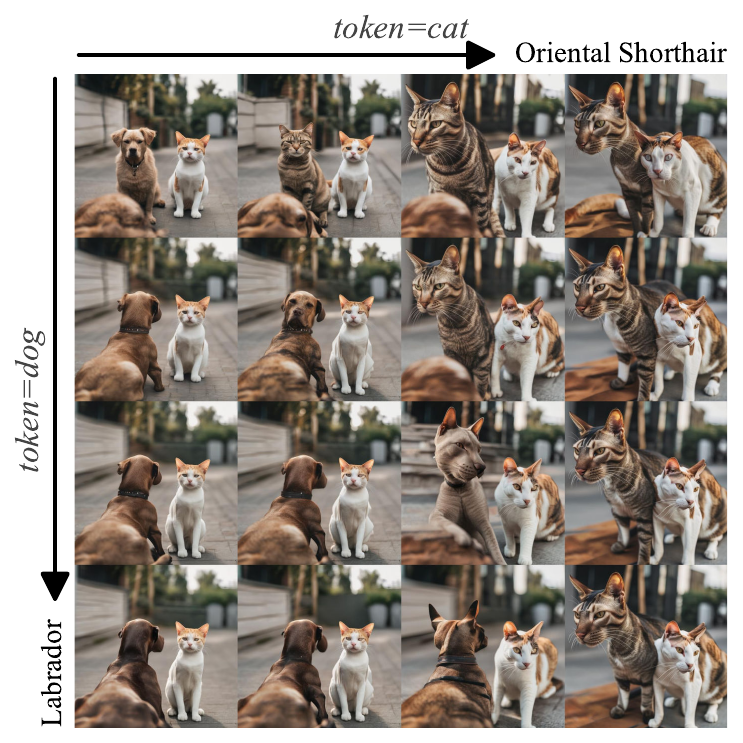}
    \end{subfigure}
    \hfill
    \begin{subfigure}[b]{0.49\textwidth}
        \centering
        \includegraphics[width=.8\textwidth]{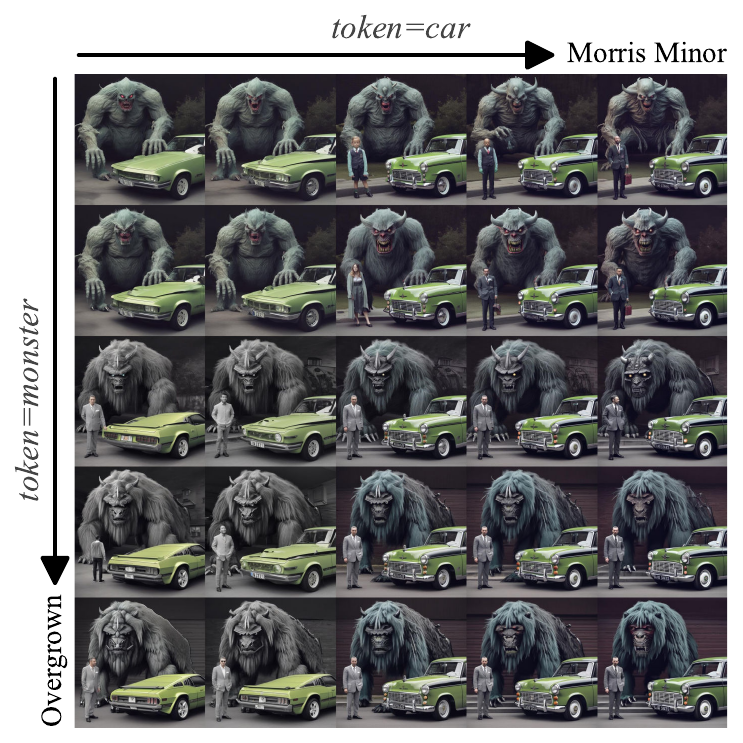} 
    \end{subfigure}
    \caption{        \label{fig:failure_composition}
    Applying latent directions discovered by \ours{} does not always allow for a token-level composability, enabling independent steering of distinct concepts. 
    }
    \label{fig:composability_failure}
\end{figure}

Moreover, while our method facilitates token-level steering, the contextual dependency of text embeddings can result in partial semantic leaking. Figure \ref{fig:failure_composition} illustrates instances where independent subject steering is compromised, leading to unintended variations in non-target subjects when the magnitude of the steering vector is high.

\label{app:failure}

\section{Equal Computation Evaluation}
\label{app:equal_compute}

To provide a fair assessment of efficiency, we benchmarked \ours{} against SliderSpace under an equal computational budget (utilizing 4 NVIDIA RTX 5000 24GB GPUs, for around 75 minutes). For this setup, we drastically reduced the number of sampled gradients for \ours{} to approximately $2,500$. As presented in Table~\ref{tab:equal_compute}, even under this constrained setup, \ours{} maintains highly competitive performance, outperforming SliderSpace in terms of intra-class diversity (DreamSim) on most concepts, while maintaining comparable or better semantic consistency (CLIP Distance). 

Furthermore, it is crucial to note a fundamental difference in how these methods scale: SliderSpace's computational cost increases linearly with each additional direction discovered, whereas \ours{} maps the entire semantic subspace and extracts all principal directions simultaneously from a single set of gradients.

\begin{table}[h]
\centering
\caption{Comparison of ELROND and SliderSpace under an equal computational budget. ELROND maintains competitive diversity and text alignment even when gradient sampling is drastically reduced.}
\label{tab:equal_compute}
\begin{tabular}{llcc}
\toprule
\textbf{Concept} & \textbf{Method} & \textbf{DreamSim} ($\uparrow$) & \textbf{CLIP} ($\uparrow$) \\
\midrule
Car     & ELROND      & 0.4814 & 0.2571 \\
        & SliderSpace & \textbf{0.5273} & 0.2511 \\ \midrule
Cat     & ELROND      & \textbf{0.3986} & 0.2693 \\
        & SliderSpace & 0.2926 & 0.2756 \\ \midrule
Dog     & ELROND      & \textbf{0.5151} & 0.2648 \\
        & SliderSpace & 0.5118 & 0.2687 \\ \midrule
Monster & ELROND      & \textbf{0.5936} & 0.2471 \\
        & SliderSpace & 0.5257 & 0.2563 \\ \midrule
Person  & ELROND      & \textbf{0.6167} & 0.2410 \\
        & SliderSpace & 0.5631 & 0.2498 \\
\bottomrule
\end{tabular}
\end{table}

\section{Discovered subspaces and SAE features analysis}
\label{app:details}

In this section, we provide a quantitative analysis of the semantic subspaces discovered by \ours{}. We compare the complexity and richness of the representations learned by the SDXL-DMD model across two decomposition methods: Principal Component Analysis (PCA) and Sparse Autoencoders (SAE).

\begin{table}[h]
        \centering
        \caption{Number of principal components found and non-dead features in the trained SAEs. All SAE have exactly 2048 features total.}
        \label{tab:pca_complexity}
        \begin{tabular}{lcc}
        \toprule
        \textbf{Concept} & \textbf{PCA} & \textbf{SAE} \\
        \midrule
        Car     & 95 & 1880 \\
        Dog     & 65 & 1601 \\
        Cat     & 77 & 1635 \\
        Person  & 68 & 1854 \\
        Monster & 74 & 1425 \\
        \bottomrule
        \end{tabular}
\end{table}

\section{Generalization to Alternative Distillation Frameworks}
\label{app:sdxl_turbo}

To verify whether our method generalizes across different distillation frameworks, we evaluated \ours{}'s performance on SDXL-Turbo. For this experiment, we utilized the PCA variant of our decomposition method and transferred the semantic directions discovered directly from the base SDXL teacher model.


\begin{table}[h]
\centering
\caption{\textbf{FID evaluation of ELROND on SDXL-Turbo.}
Lower is better. Applying ELROND consistently reduces FID across all evaluated
concepts, indicating improved agreement with the target image distribution.}
\label{tab:sdxl_turbo}
\begin{tabular}{lcc}
\toprule
\textbf{Concept} & \textbf{SDXL-Turbo} & \textbf{ELROND} \\
\midrule
Car      & 67.36  & \textbf{55.50} \\
Cat      & 70.63  & \textbf{65.19} \\
Dog      & 85.78  & \textbf{60.58} \\
Monster  & 276.73 & \textbf{245.58} \\
Person   & 186.76 & \textbf{118.43} \\
\bottomrule
\end{tabular}
\end{table}

As demonstrated in Table~\ref{tab:sdxl_turbo}, \ours{} consistently yields substantial improvements in both overall generation quality (lower FID) and intra-class diversity (higher DreamSim distance). Simultaneously, it maintains comparable or slightly improved text-to-image alignment (CLIP Distance) relative to the unmodulated base model. These findings strongly align with our primary results on SDXL-DMD, confirming that \ours{} effectively mitigates mode collapse and reintroduces lost semantic diversity across diverse distillation architectures.

\section{Loss Functions for Semantic Gradient Calculation}
\label{app:loss_ablation}

A critical component of our method is the calculation of the semantic gradient $g_{i,j}$ between a source image $x^{(i)}$ and a target image $x^{(j)}$. This gradient represents the direction in the embedding space that shifts the generation from the source state toward the target state. The choice of the loss function $\mathcal{L}_{sem}$ used to compute this gradient significantly influences the quality and steerability of the resulting semantic directions. In this section, we evaluate four different loss objective formulations to determine the optimal approach.

To assess the efficacy of different loss functions, we constructed a dataset of 200 image pairs sharing the same semantic concept (e.g., pairs of different "cats" or "cars"). For each pair $(x_{src}, x_{tgt})$, we computed the gradient $\nabla_{e_{tok}} \mathcal{L}(x_{src}, x_{tgt})$ using a specific loss function. As different loss functions naturally produce gradients of different scales, we normalize those gradients. We then initiated the reverse diffusion process using the fixed noise of the source image $x_{src}$, but intervened in the text embedding space by injecting the calculated gradient. 

The resulting generated image $x_{gen}$ represents the model's attempt to transform $x_{src}$ into $x_{tgt}$ using the semantic information encoded in the gradient. We evaluated the alignment between $x_{gen}$ and the target image $x_{tgt}$ using three metrics: DreamSim distance~\cite{fu2023dreamsim} (perceptual similarity), CLIP similarity (semantic alignment), and DINO similarity (structural/object alignment).
We compared the following four loss formulations:

\begin{enumerate}
    \item \textbf{Mean Squared Error (MSE):} The standard reconstruction loss calculated directly between the predicted denoised latent $\hat{z}_0$ of the source and the latent representation of the target.
    \begin{equation}
        \mathcal{L}_{\text{MSE}} = \| \hat{z}_0^{(src)} - z_0^{(tgt)} \|_2^2
    \end{equation}
    \item \textbf{Aligned MSE:} A variant of MSE designed to mitigate spatial misalignment between the source and target objects. We utilized DAAM~\cite{tang2023daam} to generate attention masks $M_{src}$ and $M_{tgt}$ corresponding to the concept token. We then spatially aligned the features of the target to the source based on the centroids of these masks before computing the MSE. This loss aims to penalize semantic differences while being invariant to object position.
    \item \textbf{Chamfer Distance:} Treating the feature maps of the source and target latents as sets of feature vectors, we computed the Chamfer distance. This permutation-invariant metric measures the similarity between the two sets of features without enforcing strict spatial correspondence.
    \item \textbf{Maximum Mean Discrepancy (MMD):} Similar to Chamfer distance, we treated the feature maps as distributions and minimized the MMD kernel distance between the source and target distributions.
\end{enumerate}

The quantitative results are summarized in Table~\ref{tab:loss_metrics}. We observe that the standard \textbf{MSE} loss consistently outperforms other formulations across all three metrics. It achieves the lowest DreamSim distance ($0.4114$) and the highest CLIP ($0.8037$) and DINO ($0.5073$) similarities.

While \textit{Aligned MSE} performs comparably, the overhead of computing attention maps via DAAM does not yield a performance benefit over standard MSE. The distribution-based losses (\textit{Chamfer} and \textit{MMD}) performed significantly worse, suggesting that preserving the specific spatial structure of the latent code is crucial for extracting accurate semantic directions in the diffusion embedding space. Based on these findings, we employ standard MSE for all gradient calculations in our main method.

\begin{table}[h]
    \centering
    \caption{Comparison of alignment metrics across different loss functions. The generated images steered by gradients derived from MSE loss exhibit the highest similarity to the target images across perceptual (DreamSim), semantic (CLIP), and structural (DINO) metrics.}
    \label{tab:loss_metrics}
    \begin{tabular}{l c c c}
        \toprule
        \textbf{Loss} & \textbf{DreamSim Dist.} $\downarrow$ & \textbf{CLIP Sim.} $\uparrow$ & \textbf{DINO Sim.} $\uparrow$ \\
        \midrule
        \textbf{MSE (Ours)} & \textbf{0.4114} & \textbf{0.8037} & \textbf{0.5073} \\
        Aligned MSE & 0.4123 & 0.8021 & 0.5047 \\
        Chamfer Distance & 0.4439 & 0.7837 & 0.4359 \\
        MMD & 0.4533 & 0.7707 & 0.3928 \\
        \bottomrule
    \end{tabular}
\end{table}

\section{User study}
\label{app:user_study} 
\begin{figure}[h]
    \centering
    \includegraphics[width=\textwidth]{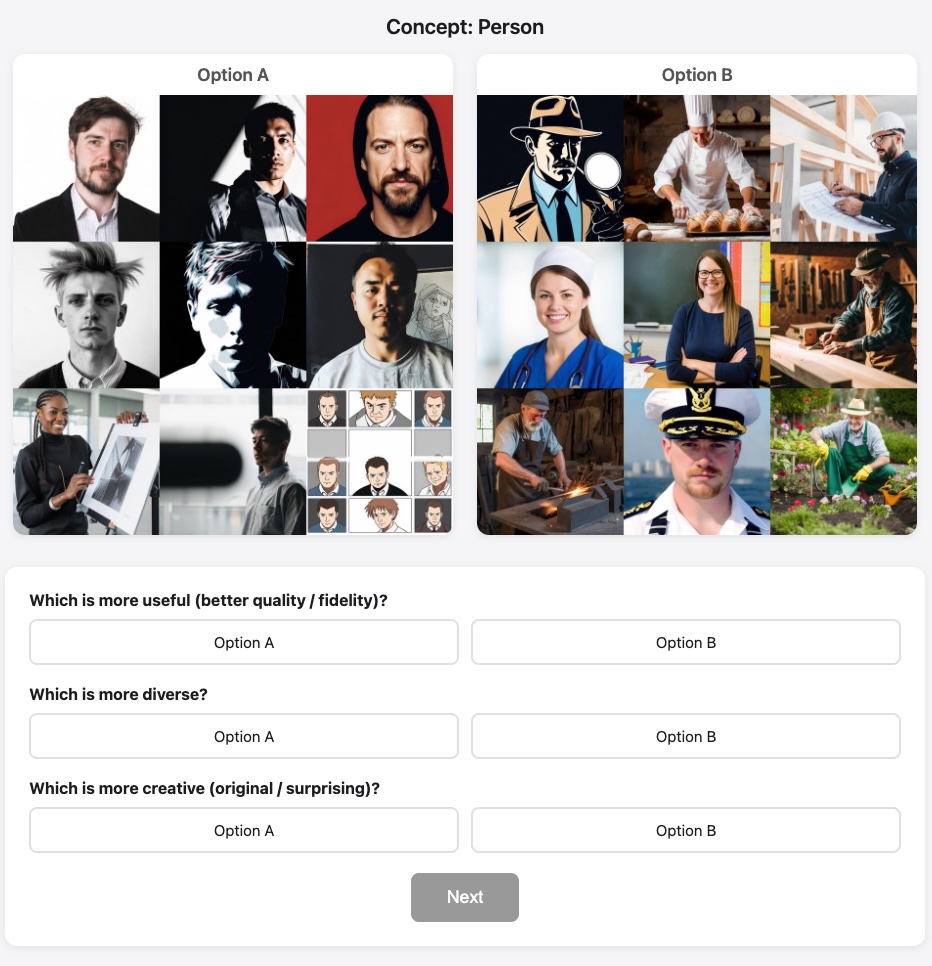}
    \caption{A screenshot of the user study questionnaire, illustrating the side-by-side image collage comparison and the specific questions targeted at usefulness, diversity, and creativity.}
    \label{fig:user_study_ui}
\end{figure}

To thoroughly evaluate the generative capabilities decomposed by our framework, we conducted a comprehensive user study. To that end, we developed a web-based questionnaire to facilitate side-by-side image comparisons. At the beginning of the study, participants were given the following instructions:
\begin{enumerate}
    \item You will see 25 pairs of generated image collages side by side.
    \item For each pair, both collages represent the same concept.
    \item For each criterion, pick the option (A or B) that you think is better.
    \item Focus solely on the properties of the main concept in the images (e.g. Dog/Car) and try to ignore style, background or other details.
    \item The three criteria are: Usefulness (quality/fidelity), Diversity (variety of outputs), and Creativity (originality/surprise).
\end{enumerate}
Figure~\ref{fig:user_study_ui} provides a screenshot of the evaluation interface presented to the users.

\paragraph{Protocol.}
Each evaluator is presented with $25$ randomly sampled pairs of collages. Every pair shows two image grids generated for the same concept: one produced by \textsc{Elrond}\,(PCA), and one produced by a
randomly selected competing method drawn uniformly from $\{\text{SDXL},\text{SDXL--DMD},\text{SliderSpace}\}$.
The five concepts considered are \emph{car}, \emph{cat}, \emph{dog}, \emph{monster}, and \emph{person}, and for each (method, concept) cell we pre-render three independent collages from which the interface
samples uniformly. The side on which \textsc{Elrond} is shown is randomised per pair so that evaluators cannot rely on positional cues, and method identities are hidden behind neutral ``Option~A'' /
``Option~B'' labels. For each pair the evaluator answers three forced-choice questions, one per criterion:
\begin{itemize}
    \item \textbf{Usefulness} -- which option has higher quality and better fidelity to the concept;
    \item \textbf{Diversity} -- which option exhibits a wider variety of distinct outputs;
    \item \textbf{Creativity} -- which option is more original or surprising.
\end{itemize}


\paragraph{Participants.}
We collected responses from $40$ evaluators, yielding a total of
$764$ submitted comparisons (mean $19.1$, median $25$ pairs per
evaluator; $28$ out of $40$ evaluators completed the full set of $25$
pairs, the remainder dropped out partway through). No
demographic information was collected and participation was
voluntary.



\section{Composability and Subject-Specific Control}
\label{app:composability}

\begin{figure}[h]
    \centering
    \begin{minipage}{0.49\linewidth}
        \centering
        \includegraphics[width=\linewidth]{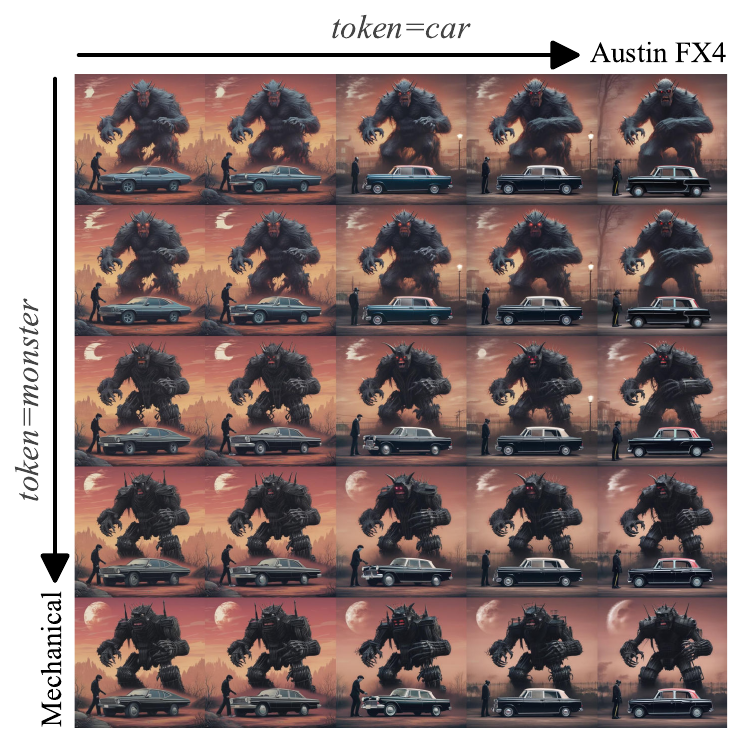}
    \end{minipage}
    \hfill
    \begin{minipage}{0.49\linewidth}
        \centering
        \includegraphics[width=\linewidth]{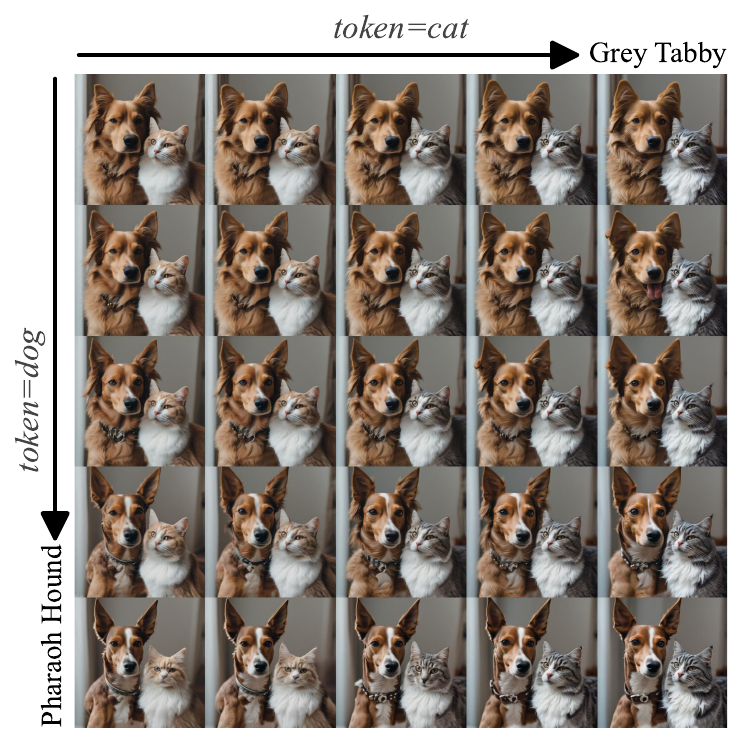}
    \end{minipage}

    \vspace{0.5em}

    \begin{minipage}{0.49\linewidth}
        \centering
        \includegraphics[width=\linewidth]{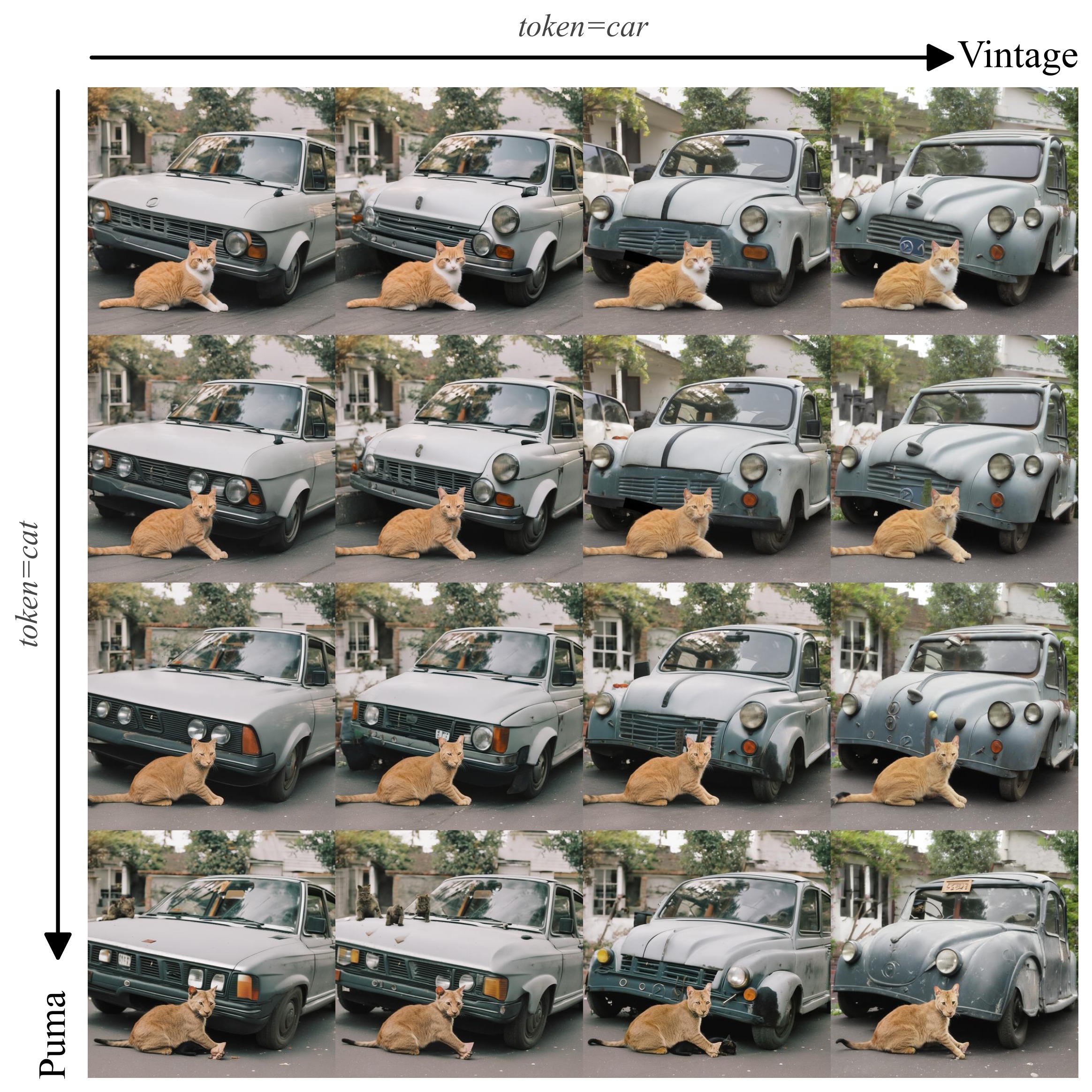}
    \end{minipage}
    \hfill
    \begin{minipage}{0.49\linewidth}
        \centering
        \includegraphics[width=\linewidth]{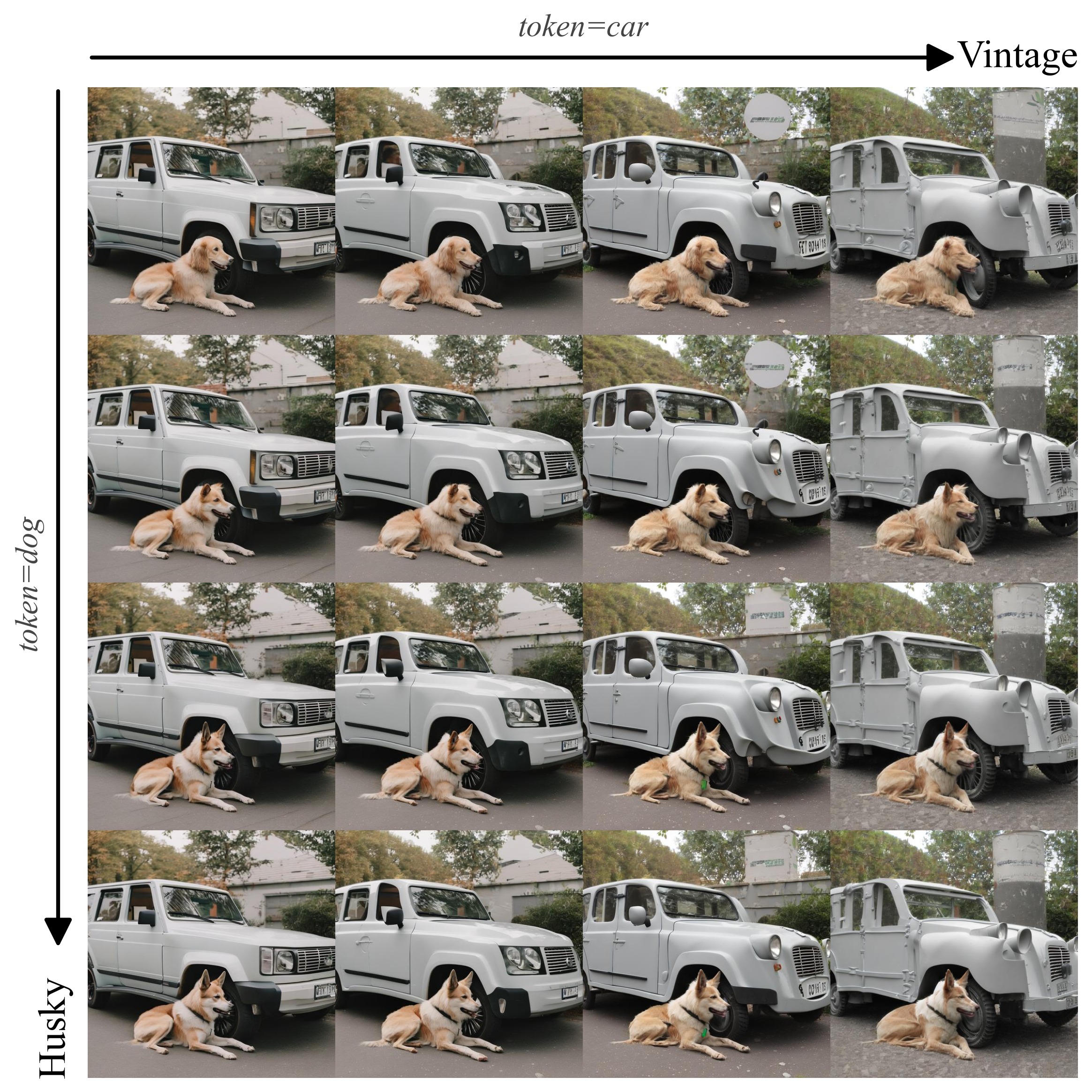}
    \end{minipage}
    \caption{Latent directions discovered by \ours{} allow for subject-specific
    control.}
    \label{fig:editing_dog_cat}
\end{figure}

A limitation of methods operating on the whole model, such as SliderSpace, is the difficulty of isolating interventions to specific subjects within a scene: because SliderSpace learns global low-rank adapters, applying a direction often affects the entire image. 
Our method instead operates on the embeddings of individual tokens, enabling token-level composability, similarly to what was observed by~\citet{garibi2025tokenverse}. Using directions discovered for \textit{Monster} and \textit{Car} in SDXL, we construct prompts containing both subjects and apply distinct directions to the corresponding tokens simultaneously. Following~\citet{gandikota2025sliderspace}, we skip the steering for the first $5$ of $50$ denoising steps, as the structural layout is determined early. Fig.~\ref{fig:editing_dog_cat} shows that our method disentangles the control of spatially distinct subjects.

\section{Additional metrics and visual comparison}

In this section, we expand our quantitative evaluation by analyzing the trade-off between the quality and diversity of the generated images using Precision and Recall metrics~\citep{kynkaanniemi2019improved}. Here, Precision measures the fidelity of the generated samples (i.e., how much they reside within the reference manifold of real data), while Recall quantifies the diversity coverage (i.e., how much of the reference manifold is captured by the generator).

\begin{table}[h]
    \centering
    \caption{\textbf{Precision} ($\uparrow$), measuring how faithful the generated
    samples are to the reference distribution.}
    \label{tab:precision_results}
    \small
    \begin{tabular}{lccccc}
        \toprule
        \textbf{Concept}
        & \makecell{\textbf{\ours{} SAE} \\ \textbf{(SDXL)}}
        & \makecell{\textbf{\ours{} SAE} \\ \textbf{(SDXL-DMD)}}
        & \makecell{\textbf{\ours{} PCA} \\ \textbf{(SDXL-DMD)}}
        & \textbf{SDXL-DMD}
        & \textbf{SliderSpace} \\
        \midrule
        Car     & 0.1492 & 0.1478 & 0.1331 & 0.1622 & \textbf{0.1673} \\
        Cat     & 0.1410 & 0.1331 & 0.0979 & \textbf{0.2306} & 0.1349 \\
        Dog     & 0.1154 & 0.1105 & 0.0847 & 0.1431 & \textbf{0.1901} \\
        Monster & \textbf{0.0710} & 0.0682 & 0.0585 & 0.0559 & 0.0649 \\
        Person  & 0.0874 & 0.0808 & 0.0808 & \textbf{0.1169} & 0.1019 \\
        \bottomrule
    \end{tabular}
\end{table}

\begin{table}[h]
    \centering
    \caption{\textbf{Recall} ($\uparrow$), measuring how much of the reference
    distribution the generated samples cover.}
    \label{tab:recall_results}
    \small
    \begin{tabular}{lccccc}
        \toprule
        \textbf{Concept}
        & \makecell{\textbf{\ours{} SAE} \\ \textbf{(SDXL)}}
        & \makecell{\textbf{\ours{} SAE} \\ \textbf{(SDXL-DMD)}}
        & \makecell{\textbf{\ours{} PCA} \\ \textbf{(SDXL-DMD)}}
        & \textbf{SDXL-DMD}
        & \textbf{SliderSpace} \\
        \midrule
        Car     & \textbf{0.5179} & 0.4782 & 0.4852 & 0.1961 & 0.4672 \\
        Cat     & 0.5240 & 0.4856 & \textbf{0.6469} & 0.1243 & 0.4318 \\
        Dog     & 0.6616 & 0.6123 & \textbf{0.7218} & 0.2059 & 0.5761 \\
        Monster & 0.4313 & 0.4793 & \textbf{0.5432} & 0.1139 & 0.4463 \\
        Person  & 0.5318 & \textbf{0.5598} & 0.5013 & 0.1296 & 0.3936 \\
        \bottomrule
    \end{tabular}
\end{table}

\paragraph{Quantifying Mode Collapse and Restoration}
The severity of mode collapse in the distilled SDXL-DMD model is explicitly captured by its low Recall scores in Table \ref{tab:recall_results}. For instance, the base SDXL-DMD achieves a Recall of only 0.2059 for the "Dog" concept, indicating that it generates a very narrow subset of the possible visual variations found in the teacher model. \ours{} successfully mitigates this collapse, drastically increasing Recall across all concepts (e.g., improving "Dog" Recall to 0.7218 with PCA), thereby verifying that our method effectively restores the intrinsic diversity of the generative process.

\paragraph{Comparison of Decomposition Methods: SAE vs. PCA}
When comparing the two decomposition strategies, a distinct trade-off between geometric coverage and semantic validity emerges. The Principal Component Analysis (PCA) variant consistently yields the highest Recall scores, such as 0.7218 for "Dog" compared to 0.6123 for the Sparse Autoencoder (SAE). This outcome is expected, as PCA maximizes variance and spans the largest possible linear subspace, thereby ensuring broadly diverse outputs. Conversely, the SAE variant consistently achieves higher Precision than PCA -- for example, 0.1105 versus 0.0847 for "Dog". This suggests that while PCA covers a larger volume of the latent space, it may inadvertently include "noisier" or less semantically coherent directions. In contrast, the SAE adapts to the non-uniform density of the semantic gradients. By minimizing expected reconstruction error, the SAE naturally allocates a higher density of feature vectors to regions of the manifold where the probability density of gradients is high. Consequently, sampling from the SAE dictionary effectively approximates the true underlying density of the concept, resulting in generated samples that are more faithful to the reference manifold and achieve consistently higher Precision scores (e.g., 0.1105 versus 0.0847 for "Dog").

\paragraph{Comparison with SliderSpace}
While SliderSpace achieves high Precision scores (e.g., 0.1901 for "Dog"), likely due to its reliance on stable, external CLIP-guided directions, it lags behind our approach in Recall (e.g., 0.5761 vs. 0.7218). This confirms our hypothesis that deriving directions from the model's own gradients is better suited for uncovering the full, latent diversity of the model than when constrained by external proxies.

For direct visual comparison in the following plots, we present random samples from the original models (SDXL and SDXL-DMD) and version augmented with directions from different decomposition techniques.

\begin{figure}[h] 
    \centering
    \includegraphics[width=.8\linewidth]{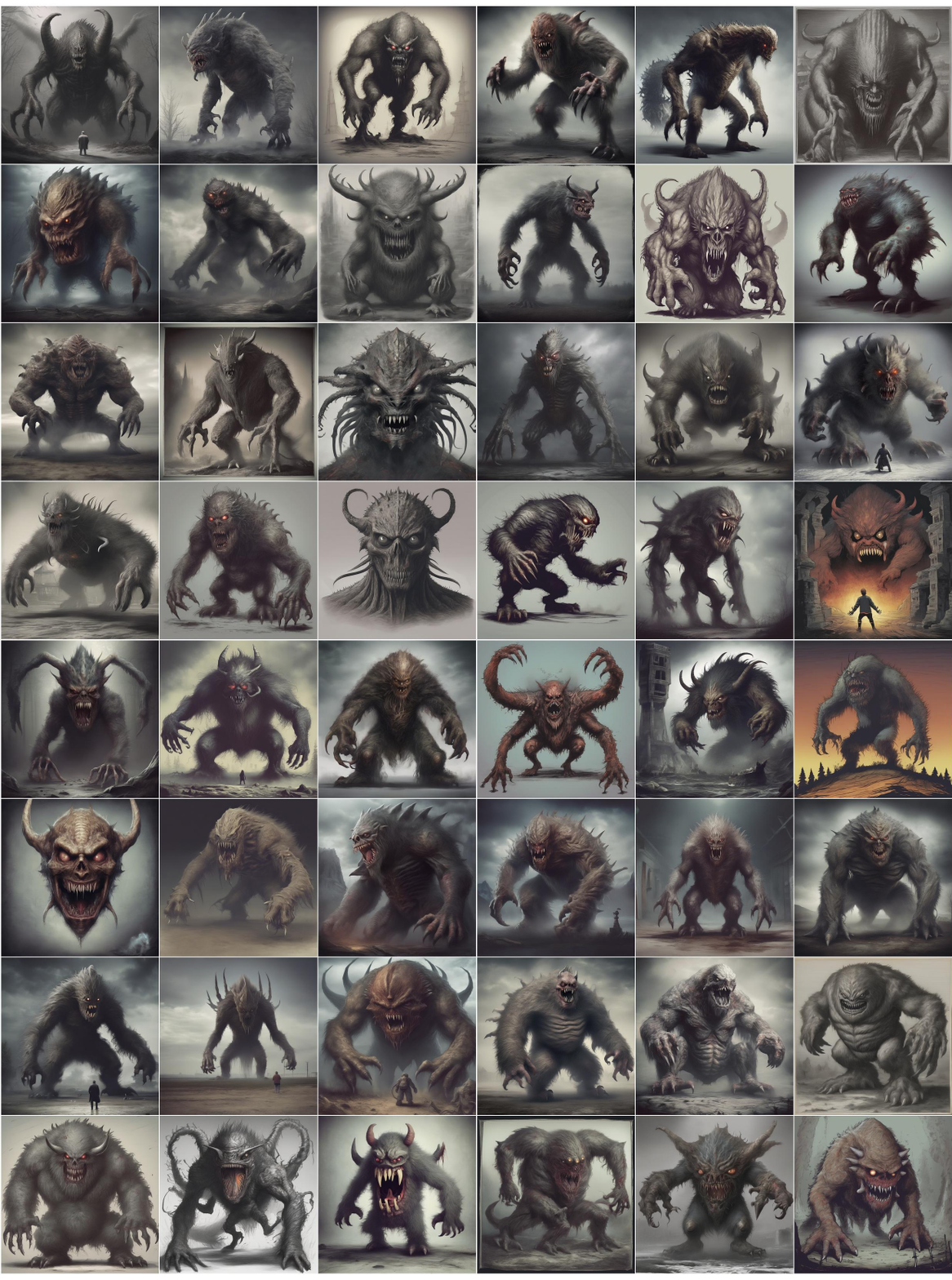}
    \caption{Uncurated samples generated by the teacher model (SDXL) for the concept "monster"
    }
    \label{fig:sdxl}
\end{figure}

\begin{figure}[h] 
    \centering
    \includegraphics[width=.8\linewidth]{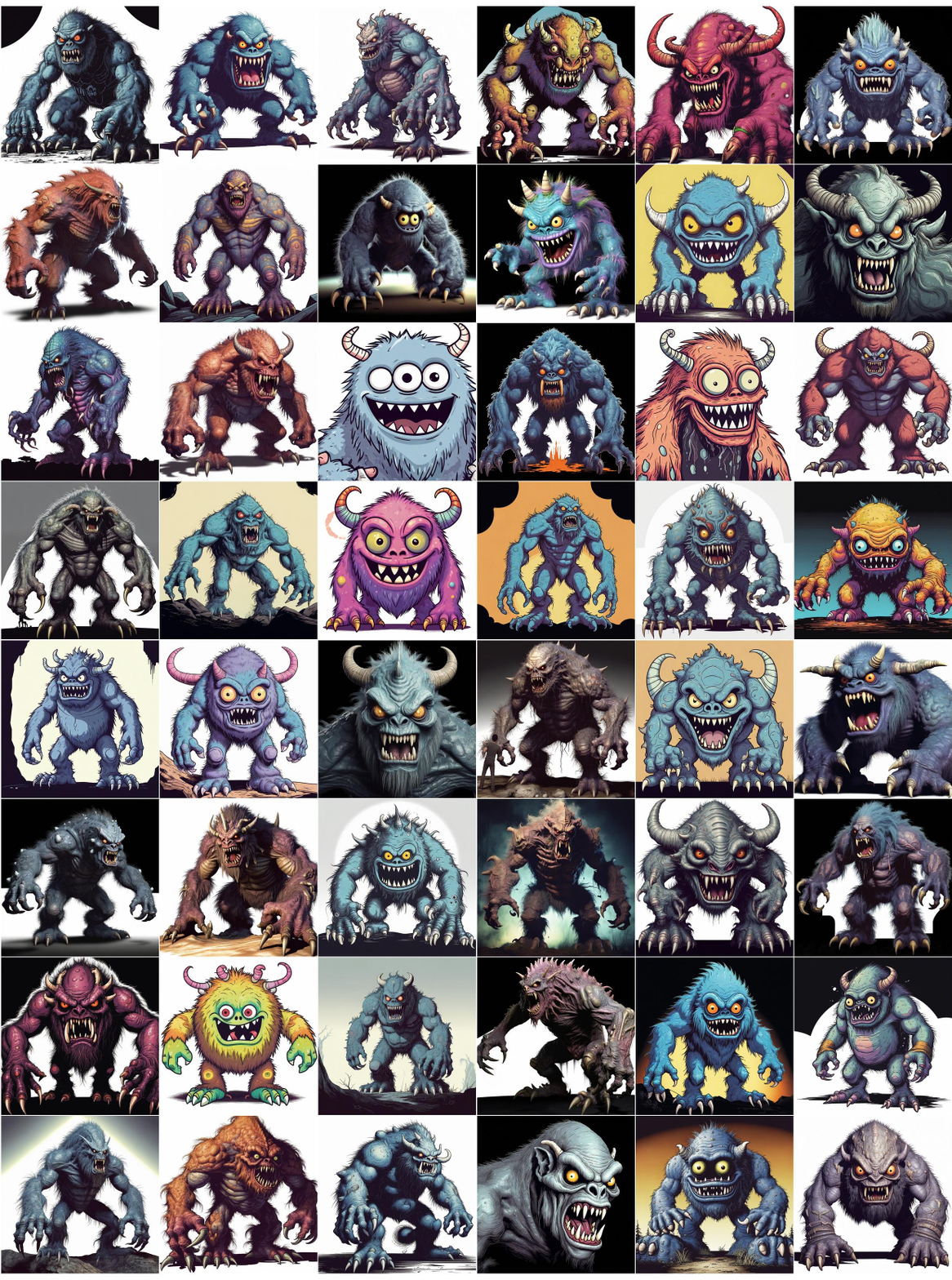}
    \caption{Uncurated samples generated by the distilled model (SDXL-DMD) for the concept "monster", illustrating the baseline performance.
    }
    \label{fig:collage_sdxl_dmd}
\end{figure}

\begin{figure}[h] 
    \centering
    \includegraphics[width=.8\linewidth]{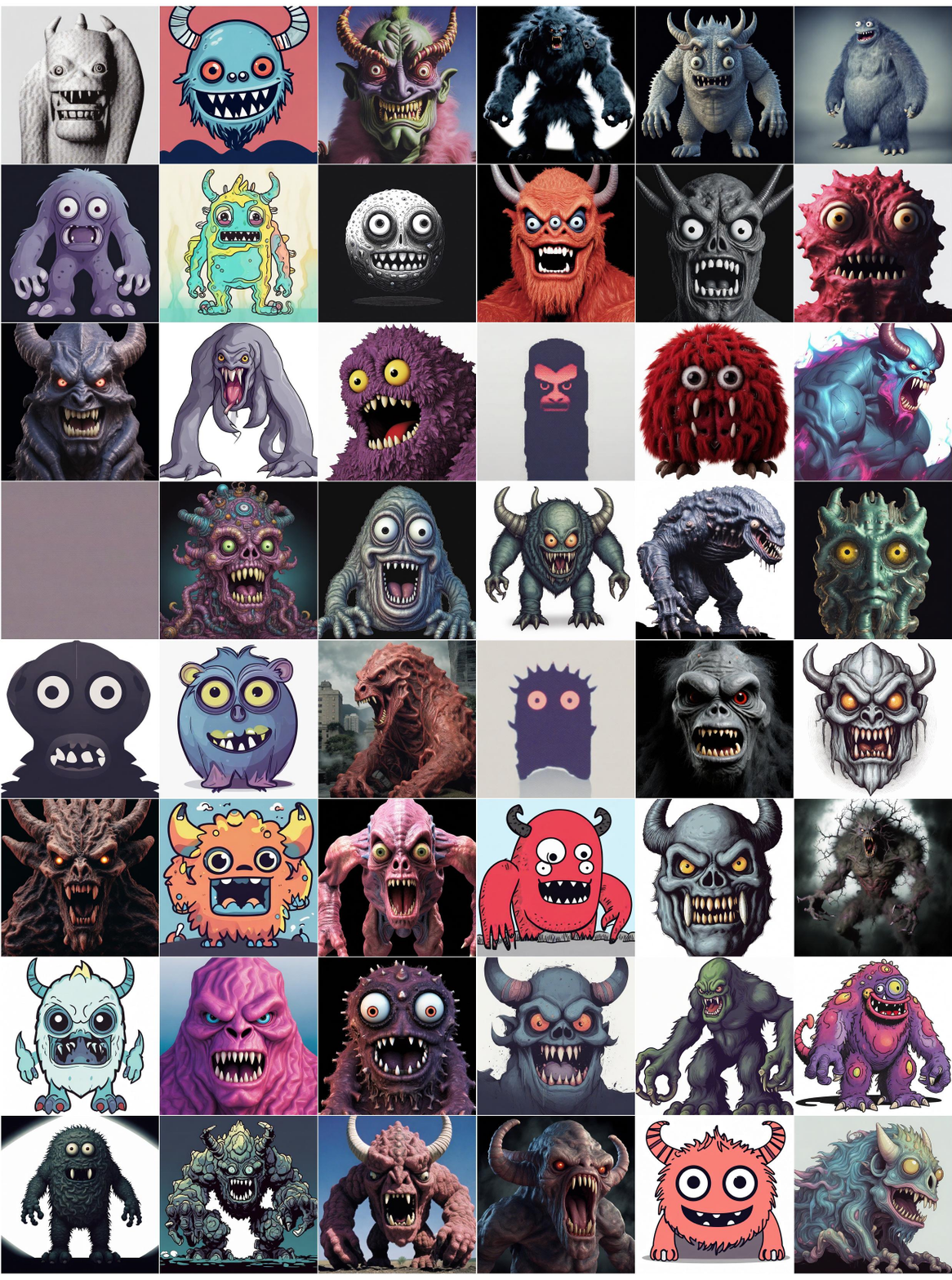}
    \caption{Samples generated by SDXL-DMD when modulated using the \textbf{SliderSpace} method.
    }
    \label{fig:collage_sliders}
\end{figure}

\begin{figure}[h] 
    \centering
    \includegraphics[width=.8\linewidth]{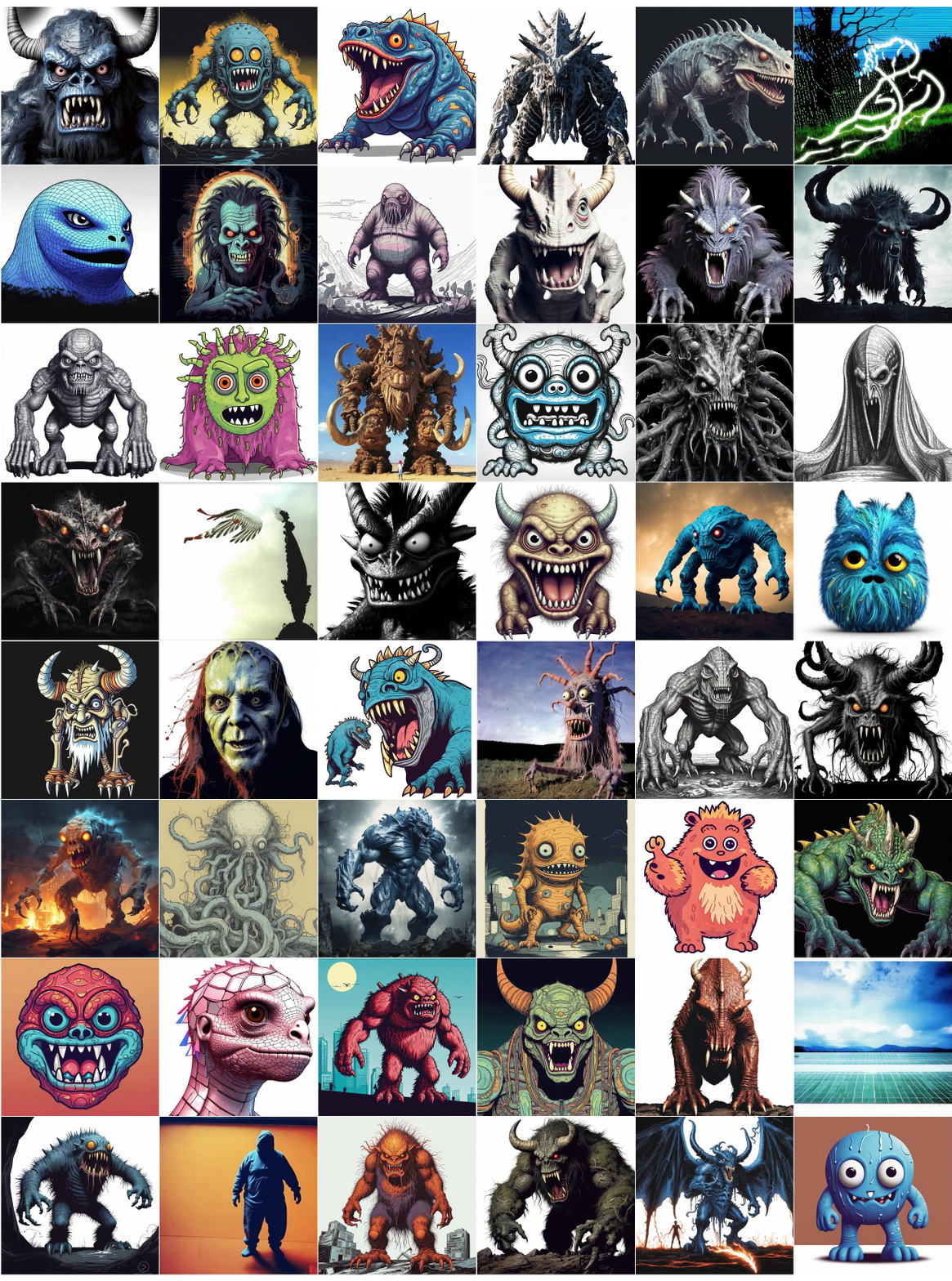}
    \caption{Samples generated by SDXL-DMD modulated with latent directions discovered by \textbf{\ours{} (PCA)} within the distilled model itself.
    }
    \label{fig:collage_elrond_pca}
\end{figure}

\begin{figure}[h] 
    \centering
    \includegraphics[width=.8\linewidth]{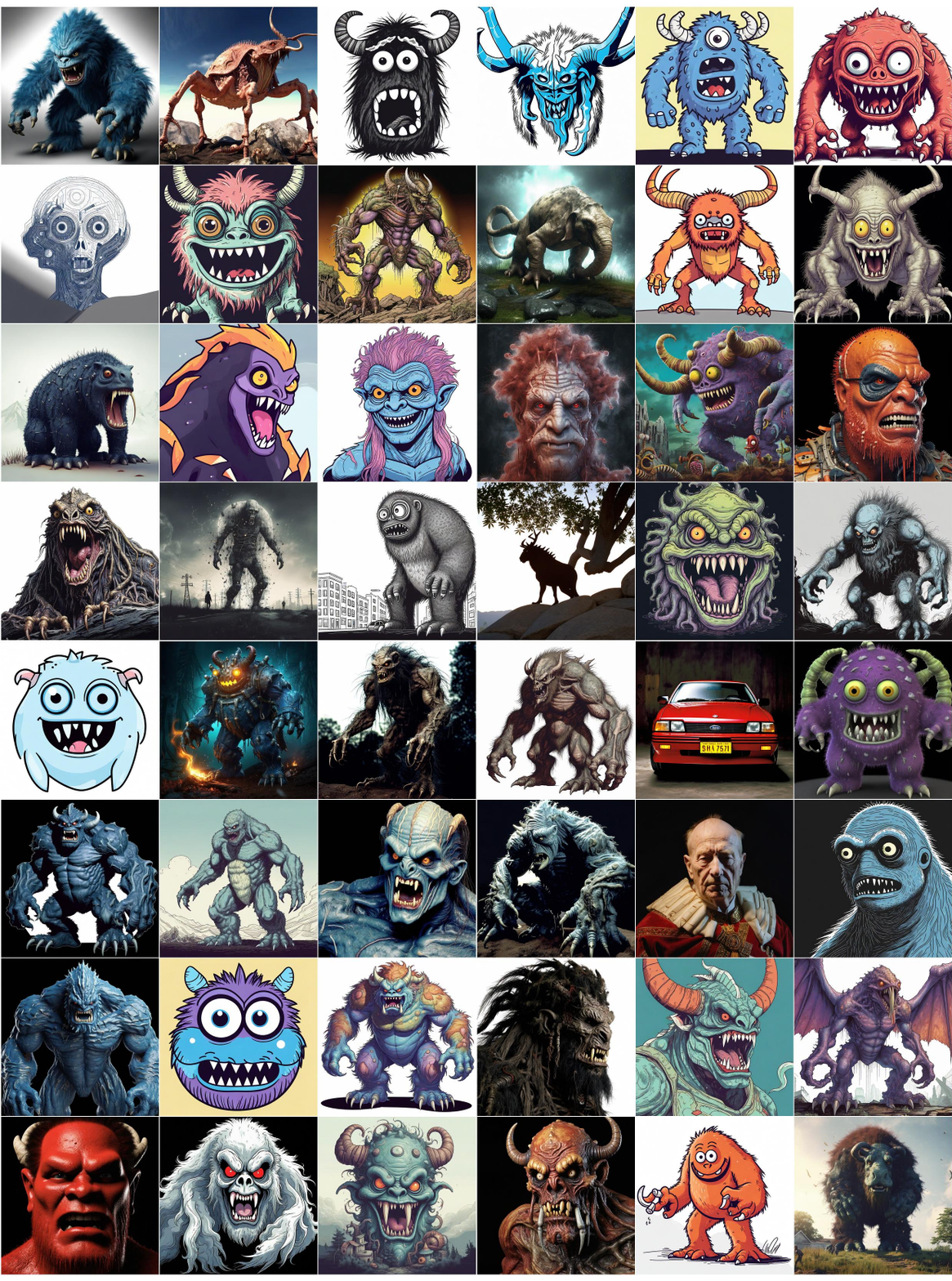}
    \caption{Samples generated by SDXL-DMD modulated with latent directions discovered by \textbf{\ours{} (SAE)} within the distilled model itself.
    }
    \label{fig:collage_elrond_sae}
\end{figure}

\begin{figure}[h] 
    \centering
    \includegraphics[width=.8\linewidth]{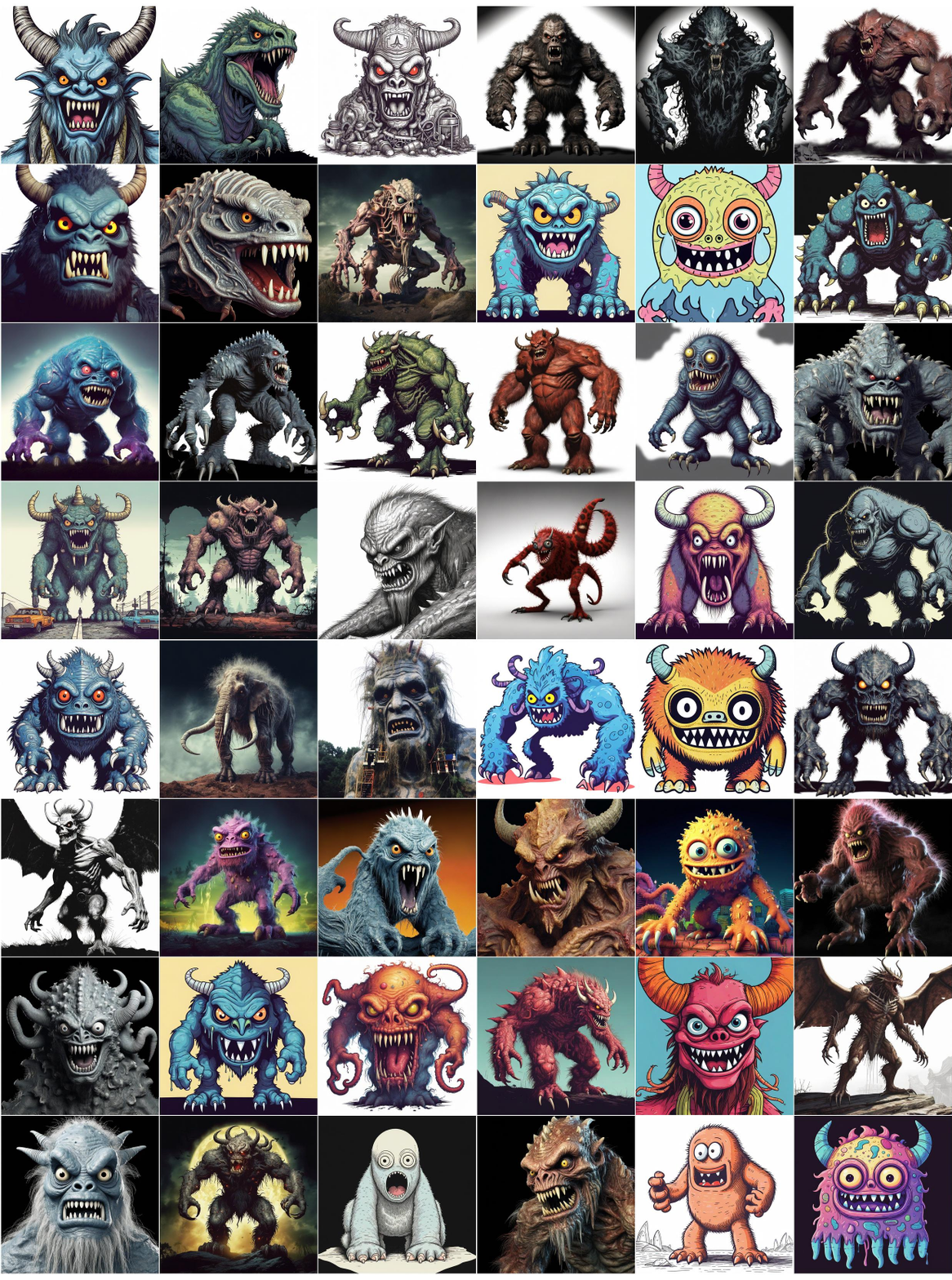}
    \caption{Samples generated by SDXL-DMD modulated with latent directions discovered by \textbf{\ours{} (SAE)} within the \textit{teacher} model (SDXL).
    }
    \label{fig:collage_elrond_sae_sdxl}
\end{figure}

\end{document}